\documentclass{article}

\usepackage{colt10e}
\usepackage{times}
\usepackage{amsfonts}
\usepackage{amsmath}
\usepackage[psamsfonts]{amssymb}
\usepackage{latexsym}
\usepackage{color}
\usepackage{graphics}
\usepackage{enumerate}
\usepackage{amstext}
\usepackage{url}
\usepackage{epsfig}

\DeclareMathOperator*{\argmin}{\rm argmin}

\newtheorem{defi}{Definition}

\newcommand{\R}{\mathbb{R}}

\newenvironment{proof*}{\noindent{\bf Proof:}}{}
\newcommand{\ignore}[1]{}
\newcommand{\bs}{\boldsymbol}
\newcommand{\vv}[1]{{\mathbf{#1}}}
\newcommand{\poly}{\mathop {\rm poly}}
\newcommand{\var}{{\mathop {\rm Var}}}

\title{Toward Learning Gaussian Mixtures with Arbitrary Separation}

\author{Mikhail Belkin\\
Ohio State University \\
Columbus, Ohio\\
\texttt{\small mbelkin@cse.ohio-state.edu}
\And Kaushik Sinha\\
Ohio State University \\
Columbus, Ohio \\
\texttt{\small sinhak@cse.ohio-state.edu}
}

\begin{document}

\maketitle

\begin{abstract}

In recent years analysis of complexity of learning Gaussian mixture models from sampled data has received
significant attention in computational machine learning and theory communities.
In this paper we present the first result showing that polynomial time learning of multidimensional Gaussian Mixture distributions is
possible when the separation between the component means is arbitrarily small.
Specifically, we present an algorithm for learning the parameters of a mixture of $k$ identical
spherical Gaussians in $n$-dimensional space with an arbitrarily small separation between the components, which is
polynomial in dimension, inverse component separation and other input
parameters for a  fixed  number of components $k$.   The algorithm uses a projection to $k$ dimensions and then a reduction to the $1$-dimensional case. 
It relies on a theoretical analysis showing that two $1$-dimensional mixtures whose densities are close in  the $L^2$ norm must have similar
means and mixing coefficients. To produce the necessary lower bound  for the $L^2$ norm in terms of the
distances between the corresponding means, we analyze the behavior of the Fourier transform
of a mixture of Gaussians  in one dimension around the origin, which turns out to be closely related
to the properties of the Vandermonde matrix obtained from the component means. Analysis of  minors of the
Vandermonde matrix  together with basic
function approximation results allows us to provide a  lower bound for the norm  of the
mixture in the  Fourier domain and hence a bound in the original space. Additionally, we present a separate argument
for reconstructing variance.

\end{abstract}


\section{Introduction}
\label{Introduction}

Mixture models, particularly Gaussian mixture models, are a widely used tool for many problems of statistical inference \cite{titter85,mclachlan00,mclachlan88,everitt81,lindsay95}.
The basic problem  is to estimate the parameters of a mixture distribution, such as the mixing
coefficients, means and variances within
some pre-specified precision from a number of sampled data points.
While the history of Gaussian mixture models goes back to~\cite{pearson1894}, in recent years the theoretical aspects  of mixture learning have attracted considerable attention  in
the theoretical computer science, starting
with the pioneering work of~\cite{dasgupta99}, who showed that a mixture of $k$ spherical
Gaussians in $n$ dimensions can be learned in time polynomial in $n$,
provided certain separation conditions between the component means (separation of order $\sqrt{n}$) are satisfied. This work has been refined and
extended in a number of recent papers. The first result from~\cite{dasgupta99} was later improved to the order of $\Omega(n^{\frac{1}{4}})$
in~\cite{dasgupta00} for spherical Gaussians and in~\cite{arora01} for general Gaussians. The separation requirement was
further reduced and made independent of $n$ to the order of $\Omega(k^{\frac{1}{4}})$ in \cite{vempala02} for spherical
Gaussians and to the order of $\Omega(\frac{k^{\frac{3}{2}}}{\epsilon^{2}})$ in~\cite{kannan05} for Logconcave distributions.
In a related work~\cite{achlioptas05} the separation requirement was reduced to $\Omega(k+\sqrt{k\log n})$. An extension of PCA
called isotropic PCA  was introduced in \cite{brubaker08}  to learn mixtures of Gaussians when any pair of
Gaussian components is separated by a hyperplane having very small overlap along the hyperplane direction (so-called "pancake layering problem").

In a slightly different direction the recent work~\cite{feldman06} made an important contribution to the subject by providing a polynomial time
algorithm  for PAC-style learning of
mixture of Gaussian distributions with arbitrary separation between the means. The authors used
a grid search over the space of parameters
to a construct a hypothesis mixture of
Gaussians that has density close to the actual mixture generating the data.
We note that the problem analyzed in~\cite{feldman06} can be viewed as
density estimation within a certain family of distributions and is different from most other work on the
subject, including our paper, which address parameter learning\footnote{Note that density estimation is
generally easier than
parameter learning since quite different configurations of parameters could conceivably lead to very
similar density functions, while similar configurations of parameters always result in similar density  functions.}.

We also note several recent papers dealing with the related problems of learning mixture of product distributions
and heavy tailed distributions. See for example, \cite{feldman08,dasgupta05,chaudhuri08a,chaudhuri08b}.

In the statistics literature, \cite{chen95} showed that optimal convergence rate of MLE estimator for finite mixture of normal distributions is $O(\sqrt{n})$, where $n$ is the sample size, if number of mixing components $k$ is known in advance and is $O(n^{-\frac{1}{4}})$ when the number of mixing components is known up to an upper bound. However, this result does not address the computational aspects, especially in high dimension.

In this paper we develop a polynomial time (for a fixed $k$) algorithm to identify the parameters of
the  mixture of $k$ identical spherical Gaussians with potentially unknown variance for an arbitrarily small separation 
between the components\footnote{We point out 
that some non-zero separation is necessary since the problem of
learning parameters without any separation assumptions at all is ill-defined.}. To the best of our knowledge this is the
first result of this  kind except for the simultaneous and independent work~\cite{kalai10},  which analyzes the case of a mixture of 
two Gaussians with arbitrary covariance matrices using the method of moments. We note that the results in~\cite{kalai10} 
and in our paper are somewhat orthogonal.  Each paper deals with a special case of the  ultimate goal (two arbitrary Gaussians in~\cite{kalai10} 
and $k$ identical spherical Gaussians with unknown variance in our case), which is to show polynomial learnability for 
a mixture with an arbitrary number of components and arbitrary variance.

All other existing algorithms for parameter estimation require minimum separation between the components to be  
an increasing function of at least one of $n$ or $k$. Our result also implies a density estimate bound
along the lines of~\cite{feldman06}.  We note, however, that we do have to pay  a price as our procedure (similarly to that in~\cite{feldman06}) is
super-exponential in $k$. Despite these limitations we believe that our paper makes a step towards understanding the
fundamental problem of polynomial learnability of Gaussian mixture distributions. 
We also think that the technique used  in the paper to obtain the lower bound may be of independent interest.

The main algorithm in our paper involves a grid search over a certain space of parameters, specifically means
and mixing coefficients of the mixture (a completely separate argument is given to estimate the variance).
By giving appropriate lower and upper bounds for the
norm of the difference of two mixture distributions in terms of their means, we show that such a grid
search is guaranteed to find a mixture with nearly correct values of the parameters.

To prove that, we need to provide a lower and upper bounds on the norm of the mixture.
A key point of our paper
is the lower bound showing that two mixtures with different means cannot produce
similar density functions. This bound is obtained by  reducing the problem to a 1-dimensional mixture
distribution and  analyzing the
behavior of the Fourier transform (closely related to the characteristic function, whose coefficients are moments
of a random variable up to multiplication by a power of the imaginary unit $i$)
of the difference between densities near zero. We use certain properties of minors of Vandermonde
matrices to show that the norm of the mixture in the Fourier domain is bounded from below.
Since the $L^2$ norm is invariant under the Fourier transform this provides a lower bound on the norm
of the mixture in the original space.

We also note the work~\cite{lindsay89}, where Vandermonde matrices appear in the analysis of mixture
distributions in the context of proving consistency of the method of moments (in fact, we rely on a result from~\cite{lindsay89} to 
provide an estimate for the variance).

Finally, our lower bound, together with an upper bound and some results from the non-parametric density estimation and
spectral projections of mixture distributions allows us to set up a grid search algorithm over the
space of parameters with the desired guarantees. 


\section{Outline of the argument}\label{sec:outline}
In this section we provide an informal outline of the argument that leads to the main result. To simplify the discussion, we will assume that the variance 
for the components is  known or estimated by using the estimation algorithm provided in  Section~\ref{sec:variance_estimation}. 
It is straightforward (but requires a lot  of  technical details) to see that all results go through if the actual variance is replaced by a 
sufficiently (polynomially) accurate estimate.

We will denote the n-dimensional Gaussian density
$\frac{1}{(\sqrt{2\pi}\sigma)^{n}}\exp\left(-\frac{\|\bs{x}-\bs{\mu}_{i}\|^{2}}{2\sigma^{2}}\right)$ by
$K(\bs{x},\bs{\mu})$, where $\bs{x},\bs{\mu}\in \R^n$ or, when appropriate, in $\R^k$. The notation $\|\cdot\|$ will always be used to represent $L^2$
norm while $d_{H}(\cdot,\cdot)$ will be used to denote the  Hausdorff distance between sets of points.
Let $p(\bs{x})=\sum_{i=1}^{k}\alpha_{i}K(\bs{x},\bs{\mu}_i)$ be a mixture of $k$
Gaussian components with the covariance matrix $\sigma^2I$ in $\mathbb{R}^n$. The goal will be to identify the means $\bs{\mu}_i$ and the mixing
coefficients $\alpha_i$ under the assumption that the minimum distance $\|\bs{\mu}_i - \bs{\mu}_j\|, i \ne j$ is bounded from below by
some given (arbitrarily
small) $d_{\min}$ and the minimum mixing weight is bounded from below by $\alpha_{\min}$. We note that while $\sigma$ can also be
estimated, we
will assume that it is known in advance to simplify the arguments. The number of
components needs to be known in advance which is in line with other work on the subject.
Our main result is an algorithm  guaranteed to produce an approximating mixture $\tilde{p}$,
whose means and mixing coefficients are all  within $\epsilon$ of their true values and whose running time is a polynomial in all parameters other than
$k$.
Input to our algorithm is $\alpha_{\min}, \sigma, k$,
$N$ points in $\R^n$ sampled from $p$ and an arbitrary small positive $\epsilon$ satisfying $\epsilon\leq\frac{d_{\min}}{2}$.
The algorithm has the following main steps.

\noindent {\bf Parameters:} $\alpha_{\min}, d_{min}, \sigma, k$.\\
{\bf Input:} $\epsilon \le \frac{d_{min}}{2}$, $N$ points  in $\R^n$ sampled from $p$.\\
{\bf Output:} $\bs{\theta^*}$, the vector of approximated means and mixing coefficients.

\textbf{Step 1.} (Reduction to $k$ dimensions).  Given a polynomial number of data points sampled from $p$ it is possible to identify
the $k$-dimensional span of the means $\bs{\mu_i}$ in $\R^n$ by using Singular Value Decomposition (see~\cite{vempala02}).
By an additional argument the problem can be reduced to analyzing a mixture of $k$ Gaussians in $\R^k$.

\textbf{Step 2.} (Construction of kernel density estimator). Using Step 1, we can assume that $n=k$. Given a sample of
$N$ points in $\R^k$, we construct a density function $p_{kde}$ using an appropriately chosen kernel density  estimator.
Given sufficiently many points, $\|p - p_{kde}\|$ can be made arbitrarily small. Note that while $p_{kde}$ is a mixture of Gaussians, it is  {\it not} a mixture of $k$ Gaussians.

\textbf{Step 3.} (Grid search).   Let ${\bs \Theta}=(\R^k)^{k}\times\R^k$ be the $k^2+k$-dimensional space of parameters (component means and mixing coefficients) to be estimated. Because of Step 1, we can assume (see Lemma \ref{lem:reduction}) $\bs{\mu}_i$s are in $\R^k$.

For any $\tilde{\bs{\theta}}=(\tilde{\bs{\mu}}_1,\tilde{\bs{\mu}}_2,\cdots,\tilde{\bs{\mu}}_k,\tilde{\bs{\alpha}})=(\tilde{\bs{m}},\tilde{\bs{\alpha}}) \in \bs{\Theta}$,  let $p(\bs{x},\tilde{\bs{\theta}})$ be the corresponding mixture distribution. Note that $\bs{\theta}=(\bs{m},\bs{\alpha})\in\bs{\Theta}$ are the true parameters. We obtain a
value $G$ (polynomial in all arguments for a fixed $k$) from Theorem~\ref{thm:main_theorem} and  take a grid $M_G$ of size $G$ in $\vv{\Theta}$. The value $\bs{\theta^*}$
is found from  a grid search according to the following equation
 \begin{equation}\label{eq:mean_selection}
\bs{\theta}^{*}=\argmin_{\tilde{\bs{\theta}} \in M_G} \left\{\|p(\bs{x},\tilde{\bs{\theta}})-p_{kde}\| \right\}
\end{equation}

We show that the means and mixing coefficients obtained by taking $\bs{\theta^*}$ are  close to the true underlying means and mixing
coefficients of $p$ with high probability. We note that our algorithm is deterministic and the uncertainty comes only from the sample
(through the SVD projection and density estimation).

%
%
%

While a somewhat different grid search algorithm was used in~\cite{feldman06},
the main novelty of our result is showing that the parameters estimated from the grid search are close to the true
underlying parameters of the mixture.  In principle, it is conceivable that two different configurations of Gaussians could give
rise to very similar mixture distributions. However, we show that this is not the case.
Specifically, and this is the theoretical core of this paper, we show that mixtures with different means/mixing
coefficients cannot be close in $L^2$ norm\footnote{Note that our notion of distance between two density functions is slightly different from the standard ones used in literature, e.g., Hellinger distance or KL divergence. However, our goal is to estimate the parameters and here we use $L^2$ norm  merely as a tool to describe that two distributions are different. } (Theorem \ref{thm:multivariate_lower_bound}) and
thus the grid search yields parameter values $\bs \theta^*$  that are close to the true values of the means and mixing coefficients.

To provide a better high-level overview of the whole proof we give a high level summary of the argument (Steps 2 and 3). 
\begin{enumerate}

\item Since we do not know the underlying probability distribution $p$ directly, we construct
$p_{kde}$, which is a proxy for $p = p(\bs{x},\bs{\theta})$. $p_{kde}$ is obtained by taking an appropriate non-parametric density estimate and, given a sufficiently large
polynomial
sample,
can be made to be arbitrarily close to $p$ in $L^2$ norm (see Lemma \ref{lem:kde_estimate}).
Thus the problem of approximating $p$ in $L^2$ norm can be replaced by
approximating $p_{kde}$.

\item The main technical part of the paper are the lower and upper bounds on the norm $\|p(\bs{x},\bs{\theta})-p(\bs{x},\tilde{\bs{\theta}})\|$ in terms of the Hausdorff distance between the  component means  (considered as sets of $k$ points) $\bs{m}$ and $\tilde{\bs{m}}$.
Specifically, in Theorem \ref{thm:multivariate_lower_bound} and Lemma \ref{thm:multivariate_upper_bound} we prove that for $\tilde{\bs{\theta}} = (\tilde{\bs{m}},\tilde{\bs{\alpha}})$
$$
d_{H}(\bs{m},\tilde{\bs{m}})\leq f(\|p(\bs{x},\bs{\theta})-p(\bs{x},\tilde{\bs{\theta}})\|)\leq h(d_{H}(\bs{m},
\tilde{\bs{m}}) + \|\bs{\alpha}  - \tilde{\bs{\alpha}}\|_1)
$$
where $f,h$ are some explicitly given increasing functions. The lower bound shows that $d_{H}(\bs{m},\tilde{\bs{m}})$  can be controlled by
making $\|p(\bs{x},\bs{\theta})-p(\bs{x},\tilde{\bs{\theta}})\|$ sufficiently small, which (assuming minimum separation $d_{min}$ between the components
of $p$) immediately
implies that each component mean of  $\bs{m}$ is close to  exactly one component mean of $\tilde{\bs{m}}$.

On the other hand, the upper bound guarantees that a search over a sufficiently fine grid in the space $\bs{\Theta}$ will produce a value $\bs{\theta}^*$, s.t.
$\|p(\bs{x},\bs{\theta})-p(\bs{x},\bs{\theta}^*)\|$ is small.

\item Once the component means $\bs{m}$ and $\tilde{\bs{m}}$ are shown to be close an argument using the Lipschitz property
of the mixture with respect to the mean locations can be used to establish that the corresponding mixing coefficient are also close (Corollary \ref{cor:mixing_weight}).

%


\end{enumerate}
We will now briefly outline the argument for the main theoretical contribution of this paper which is a lower bound on the $L^2$ norm in terms of the Hausdorff distance (Theorem~\ref{thm:multivariate_lower_bound}).
\begin{enumerate}
\item  (Minimum distance, reduction from $\R^k$  to $\R^1$) Suppose a  component mean $\bs{\mu}_i$, is separated from every estimated mean $\tilde{\bs{\mu}}_{j}$  by a distance of at least $d$,  then there exists a unit vector $\bs{v}$ in $\mathbb{R}^{k}$
such than $\forall_{i, j}$ $|\langle\bs{v}, (\tilde{\bs{\mu}}_i-\bs{\mu}_j)\rangle|\geq\frac{d}{4k^{2}}$.
In other words a certain amount of separation is preserved after an appropriate projection to one dimension. See Lemma~\ref{lem:existence_of_direction}
for a proof.

\item (Norm estimation, reduction from $\R^k$  to $\R^1$). Let $p$ and $\tilde{p}$ be the true and estimated density respectively and
let $\bs{v}$ be a unit vector in $\R^k$. $p_{\bs{v}}$ and $\tilde{p}_{\bs{v}}$ will denote the one-dimensional marginal densities obtained by integrating
$p$ and $\tilde{p}$ in the directions orthogonal to $\bs{v}$. It is easy to see that $p_{\bs{v}}$ and $\tilde{p}_{\bs{v}}$ are mixtures of
$1$-dimensional Gaussians, whose means are projections of the original means onto $\bs{v}$.
It is shown in Lemma~\ref{lem:norm_projection_bound} that
$$
\|p-\tilde{p}\|^{2}\geq\left(\frac{1}{c\sigma}\right)^{k}\|p_{\bs{v}}-\tilde{p}_{\bs{v}}\|^{2}
$$
and thus to provide a lower bound for $\|p-\tilde{p}\|$ it is sufficient to provide an analogous bound (with a different separation between the means)
in one dimension.

\item ($1$-d lower bound)
Finally, we consider a mixture $q$ of $2k$ Gaussians in one dimension, with the assumption that one of the component means is separated from the
rest of the component means by at least $t$ and that the (not necessarily positive) mixing weights exceed $\alpha_{min}$ in absolute value.
Assuming that the  means lie in an interval $[-a,a]$ we show (Theorem~\ref{thm:1d_mixture_lower_bound})
$$\|q\|^{2}\geq\alpha_{\min}^{4k}\left(\frac{t}{a^2}\right)^{Ck^{2}}$$
for some positive constant $C$ independent of $k$.\\

The proof of this result relies on analyzing the Taylor series for the Fourier transform of $q$ near zeros, which turns out to be
closely related to a  certain Vandermonde matrix.

\end{enumerate}
Combining 1 and 2 above and applying the result in 3, $q = p_{\bs{v}}-\tilde{p}_{\bs{v}}$  yields the desired lower bound for $\|p-\tilde{p}\|$.

\section{Main Results}

In this section we present our main results. First we show that we can reduce the problem in $\R^n$ to a corresponding
problem in $\R^k$
, where $n$ represents the dimension and $k$ is the number of components, at the cost of an arbitrarily small error.
Then we solve the reduced problem in $\R^k$, again allowing for only an arbitrarily small error, by establishing appropriate
lower and upper bounds of a mixture norm in $\R^k$.

\begin{lemma}[Reduction from $\R^n$ to $\R^k$]\label{lem:reduction}
Consider a mixture of $k$ n-dimensional spherical Gaussians  $p(\bs{x})=\sum_{i=1}^{k}\alpha_{i}K(\bs{x},\bs{\mu}_{i})$ where the means lie within a
cube $[-1,1]^{n}$, $\|\bs{\mu}_i-\bs{\mu}_j\| \geq d_{\min}>0, \forall_{i\neq j}$ and for all $i$,
$\alpha_i >\alpha_{\min}$. For any positive $\epsilon\leq\frac{d_{\min}}{2}$ and $\delta\in(0,1)$, given a sample of size
$\poly\left(\frac{n}{\epsilon\alpha_{\min}}\right)\cdot\log\left(\frac{1}{\delta}\right)$, with probability greater than $1-\delta$, the problem of learning the
parameters (means and mixing weights) of $p$ within
$\epsilon$ error can be reduced to learning the parameters of a k-dimensional mixture of
spherical Gaussians $p_o(\bs{x})=\sum_{i=1}^{k}\alpha_{i}K(\bs{x},\bs{\nu}_{i})$ where the means lie within a
cube $[-\sqrt{\frac{n}{k}},\sqrt{\frac{n}{k}}]^{k}$, $\|\bs{\nu}_i-\bs{\nu}_j\| > \frac{d_{\min}}{2}>0, \forall_{i\neq j}$.
However, in $\R^k$ we need to learn the means within $\frac{\epsilon}{2}$ error.
\end{lemma}
\begin{proof}
For $i=1,\ldots,k$, let $\bs{v}_i\in\mathbb{R}^{n}$ be the top $k$ right singular vectors of a data matrix of size $\poly\left(\frac{n}{\epsilon\alpha_{\min}}\right)\cdot\log\left(\frac{1}{\delta}\right)$ sampled from $p(\bs{x})$. It is well known (see \cite{vempala02})
that the space spanned by the means $\{\bs{\mu}_i\}_{i=1}^k$ remains arbitrarily close to the space spanned by $\{\bs{v}_i\}_{i=1}^k$.
In particular, with probability greater than $1-\delta$, the projected means $\{\tilde{\bs{\mu}}_i\}_{i=1}^k$ satisfy  $\|\bs{\mu}_i-\tilde{\bs{\mu}}_i\|\leq\frac{\epsilon}{2}$  for all $i$ (see Lemma \ref{lem:svd_projection}).

Note that each projected mean $\tilde{\bs{\mu}}_i\in\R^n$ can be represented by a $k$ dimensional vector
$\bs{\nu}_i$ which are the coefficients along the singular vectors $\bs{v}_j$s, that is for
all $i,~ \tilde{\bs{\mu}}_i=\sum_{j=1}^k\nu_{ij}\bs{v}_j$. Thus, for any $i\neq j ,\|\tilde{\bs{\mu}}_i-\tilde{\bs{\mu}}_j\|=\|\bs{\nu}_i-\bs{\nu}_j\|$.
Since $\|\tilde{\bs{\mu}}_i-\tilde{\bs{\mu}}_j\|\geq d_{\min}-\frac{\epsilon}{2}-\frac{\epsilon}{2}=d_{\min}-\epsilon\geq d_{\min}-\frac{d_{\min}}{2}=\frac{d_{\min}}{2}$, we have $\|\bs{\nu}_i-\bs{\nu}_j\|\geq\frac{d_{\min}}{2}$. Also note that
each $\bs{\nu}_{i}$ lie within a cube
of $[-\sqrt{\frac{n}{k}}, \sqrt{\frac{n}{k}}]^{k}$ where the axes of the cube are along the top $k$ singular vectors $\bs{v}_j$s.

Now suppose we can estimate each $\bs{\nu}_i$ by
$\tilde{\bs{\nu}}_i\in\R^k$ such that $\|\bs{\nu}_i-\tilde{\bs{\nu}}_i\|\leq\frac{\epsilon}{2}$. Again
each $\tilde{\bs{\nu}}_i$ has a corresponding representation $\hat{\bs{\mu}}_i\in\R^n$ such that $\hat{\bs{\mu}}_i=\sum_{j=1}^k\tilde{\nu}_{ij}\bs{v}_j$ and
$\|\tilde{\bs{\mu}}_i-\hat{\bs{\mu}}_i\|=\|\bs{\nu}_i-\tilde{\bs{\nu}}_i\|$. This implies
for each $i$, $\|\bs{\mu}_i-\hat{\bs{\mu}}_i\|\leq\|\bs{\mu}_i-\tilde{\bs{\mu}}_i\|+\|\tilde{\bs{\mu}}_i-\hat{\bs{\mu}}_i\|\leq\frac{\epsilon}{2}+\frac{\epsilon}{2}=\epsilon$.
\end{proof}

From here onwards we will deal with mixture of Gaussians in $\R^k$. Thus we will assume that $p_o$ denotes the true
mixture with means $\{\bs{\nu}_i\}_{i=1}^k$
while $\tilde{p}_o$ represents any other mixture in $\R^k$ with different means and mixing weights.

We first prove a lower bound for $\|p_o-\tilde{p}_o\|$.
\begin{theorem}[Lower bound in $\R^k$]\label{thm:multivariate_lower_bound}
Consider a mixture of $k$ k-dimensional spherical Gaussians $p_o(\bs{x})=\sum_{i=1}^{k}\alpha_{i}K(\bs{x},\bs{\nu}_i)$
where the means lie within a cube $[-\sqrt{\frac{n}{k}}, \sqrt{\frac{n}{k}}]^{k}$, $\|\bs{\nu}_i-\bs{\nu}_j\| \geq \frac{d_{\min}}{2}>0, \forall_{i\neq j}$ and for all $i$,$\alpha_i >\alpha_{\min}$.
Let $\tilde{p}_o(\bs{x})=\sum_{i=1}^{k}\tilde{\alpha}_i K(\bs{x},\tilde{\bs{\nu}}_{i})$ be some arbitrary mixture
such that the Hausdorff distance between the set of true means $\bs{m}$ and the estimated means $\tilde{\bs{m}}$
satisfies $d_{H}(\bs{m},\tilde{\bs{m}})\leq\frac{d_{\min}}{4}$.
Then $\|p_o-\tilde{p}_o\|^{2}\geq\left(\frac{\alpha_{\min}^4}{c\sigma}\right)^{k}\left(\frac{d_{H}(\bs{m},\tilde{\bs{m}})}{4nk^{2}}\right)^{Ck^2}$ where
  $C,c$ are some positive constants independent of $n,k$.
\end{theorem}
\begin{proof}
Consider any arbitrary $\bs{\nu}_i$ such that its closest estimate $\tilde{\bs{\nu}}_i$ from $\tilde{\bs{m}}$ is $t=\|\bs{\nu}_i-\tilde{\bs{\nu}}_i\|$. 
Note that $t\leq \frac{d_{\min}}{4}$ and all other $\bs{\nu}_j, \tilde{\bs{\nu}}_j, j\neq i$ are at a distance at least $t$ from $\bs{\nu}_i$. Lemma \ref{lem:existence_of_direction} ensures the existence of a direction $\bs{v}\in\R^k$ such that upon projecting on which $|\langle \bs{v}, (\bs{\nu}_i-\tilde{\bs{\nu}}_i)\rangle|\geq\frac{t}{4k^2}$ and all other projected means $\langle \bs{v},\bs{\nu}_j\rangle , \langle \bs{v},\tilde{\bs{\nu}}_j\rangle, j\neq i$ are at a distance at least $\frac{t}{4k^2}$ from $\langle \bs{v},\bs{\nu}_i\rangle$. Note that after projecting on $\bs{v}$, the mixture becomes a mixture of 1-dimensional Gaussians with
variance $\sigma^2$ and whose projected means lie within $[-\sqrt{n},\sqrt{n}]$. Let us denote these 1-dimensional mixtures
by $p_{\bs{v}}$ and $\tilde{p}_{\bs{v}}$ respectively. Then using Theorem \ref{thm:1d_mixture_lower_bound}
$\|p_{\bs{v}}-\tilde{p}_{\bs{v}}\|^2\geq \alpha_{\min}^{4k}\left(\frac{(t/4k^2)}{n}\right)^{Ck^2}$. Note that we
obtain $p_{\bs{v}}$ (respectively $\tilde{p}_{\bs{v}}$) by integrating $p_o$ (respectively $\tilde{p}_o$) in all $(k-1)$
orthogonal directions to $\bs{v}$. Now we need to relate $\|p_o-\tilde{p}_o\|$ and $\|p_{\bs{v}}-\tilde{p}_{\bs{v}}\|$.
This is done in Lemma \ref{lem:norm_projection_bound} to ensure that $\|p_o-\tilde{p}_o\|^2\geq \left(\frac{1}{c\sigma}\right)^{k}\|p_{\bs{v}}-\tilde{p}_{\bs{v}}\|^2$ where $c>$ is in chosen such a way that in
any arbitrary direction probability mass of each projected Gaussian on that direction becomes negligible outside the
interval of $[-c\sigma/2,c\sigma/2]$. Thus, $\|p_o-\tilde{p}_o\|^{2}\geq\left(\frac{\alpha_{\min^4}}{c\sigma}\right)^{k}\left(\frac{t}{4nk^{2}}\right)^{Ck^2}$.
Since this holds for any arbitrary $\bs{\nu}_i$, we can replace $t$ by $d_{H}(\bs{m},\tilde{\bs{m}})$.
\end{proof}

Next, we prove a straightforward upper bound for $\|p_o-\tilde{p}_o\|$.
\begin{lemma}[Upper bound in $\R^k$]\label{thm:multivariate_upper_bound} Consider a mixture of $k$, $k$-dimensional spherical Gaussians $p_o(\bs{x})=\sum_{i=1}^{k}\alpha_{i}K(\bs{x},\bs{\nu}_i)$
where the means lie within a cube $[-\sqrt{\frac{n}{k}}, \sqrt{\frac{n}{k}}]^{k}$,
$\|\bs{\nu}_i-\bs{\nu}_j\| \geq \frac{d_{\min}}{2}>0, \forall_{i\neq j}$ and for all $i$,$\alpha_i >\alpha_{\min}$.
Let $\tilde{p}_o(\bs{x})=\sum_{i=1}^{k}\tilde{\alpha}_i K(\bs{x},\tilde{\bs{\nu}}_{i})$ be some arbitrary mixture
such that the Hausdorff distance between the set of true means $\bs{m}$ and the estimated means $\tilde{\bs{m}}$
satisfies $d_{H}(\bs{m},\tilde{\bs{m}})\leq\frac{d_{\min}}{4}$. 
Then there exists a permutation $\pi:\{1,2,\ldots,k\}\rightarrow\{1,2,,\ldots,k\}$ such that
$$\|p_o-\tilde{p}_o\|\leq\frac{1}{(2\pi\sigma^2)^{k/2}}\sum_{i=1}^k\left(\sqrt{|\alpha_i-\tilde{\alpha}_{\pi(i)}|^2+\frac{d_{H}^2(\bs{m},\tilde{\bs{m}})}{\sigma^2}}\right)$$
\end{lemma}
\begin{proof}
Due to the constraint on the Hausdorff distance and constraint on the pair wise distance between the means of $\bs{m}$, there exists a permutation $\pi:\{1,2,\ldots,k\}\rightarrow\{1,2,,\ldots,k\}$ such that $\|\bs{\nu}_i-\hat{\bs{\nu}}_{\pi(i)}\|\leq d_H(\bs{m},\tilde{\bs{m}})$.
Due to one-to-one correspondence, without loss of generality we can write,\\
$\|p_o-\tilde{p}_o\|\leq\sum_{i=1}^{k}||g_{i}\|$ where $g_{i}(\bs{x})=\alpha_{i}K(\bs{x},\bs{\nu}_{i})-\tilde{\alpha}_{\pi(i)} K(\bs{x},\tilde{\bs{\nu}}_{\pi(i)})$.
Now using Lemma~\ref{lem:upper_bound}, \\ $\|g_{i}\|^{2}\leq\frac{1}{(2\pi\sigma^{2})^{k}}\left(\alpha_i^2+\tilde{\alpha}_{\pi(i)}^2-2\alpha_i\tilde{\alpha}_{\pi(i)}\exp\left(-\frac{\|\bs{\nu}_{i}-\tilde{\bs{\nu}}_{\pi(i)}\|^{2}}{2\sigma^{2}}\right)\right)$\\
$=\frac{1}{(2\pi\sigma^{2})^{k}}\left((\alpha_i-\tilde{\alpha}_{\pi(i)})^2+2\alpha_i\tilde{\alpha}_{\pi(i)}\left(1-\exp\left(-\frac{\|\bs{\nu}_{i}-\tilde{\bs{\nu}}_{\pi(i)}\|^{2}}{2\sigma^{2}}\right)\right)\right)$\\
$\leq\frac{1}{(2\pi\sigma^{2})^{k}}\left((\alpha_i-\tilde{\alpha}_{\pi(i)})^2+\frac{\alpha_i\tilde{\alpha}_{\pi(i)}\|\bs{\nu}_i-\tilde{\bs{\nu}}_{\pi(i)}\|^{2}}{\sigma^{2}}\right)$
\end{proof}

We now present our main result for learning mixture of Gaussians with arbitrary small separation.
\begin{theorem}\label{thm:main_theorem}
Consider a mixture of $k$ n-dimensional spherical Gaussians  $p(\bs{x})=\sum_{i=1}^{k}\alpha_{i}K(\bs{x},\bs{\mu}_{i})$
where the means lie within a cube $[-1,1]^{n}$, $\|\bs{\mu}_i-\bs{\mu}_j\| > d_{\min}>0, \forall_{i\neq j}$ and for all $i$,
$\alpha_i >\alpha_{\min}$. Then given any positive
$\epsilon\leq\frac{d_{\min}}{2}$ and $\delta\in(0,1)$, there exists a positive $C_1$ independent of $n$ and $k$ such that
using a sample of size
$N=\poly\left(\left(\frac{nk^2}{\epsilon}\right)^{k^3}\cdot\log^k\left(\frac{2}{\delta}\right)\right)$
and a grid $M_{G}$ of size
$G=\frac{(\alpha_{\min}^4)^k}{k^{3/2}}\left(\frac{\epsilon}{8nk^2}\right)^{C_1k^2}$,
our algorithm given by Equation \ref{eq:mean_selection} runs in time
$\frac{k^{3/2}}{(\alpha_{\min}^4\sigma)^k}\left(\frac{n^{3/2}k^{1/2}}{\epsilon}\right)^{C_1k^2}$ and provides
mean estimates which, with probability greater than $1-\delta$, are within $\epsilon$ of their
corresponding true values.
\end{theorem}
\begin{proof}
The proof has several parts.\\
\textbf{SVD projection}:
We have shown in Lemma \ref{lem:reduction} that after projecting to SVD space (using a sample of size
$\poly\left(\frac{n}{\alpha_{\min}\epsilon}\right)\cdot\log\left(\frac{2}{\delta}\right)$), we need to estimate the parameters
of the mixture in $\R^k$, $p_o(\bs{x})=\sum_{i=1}^{k}\alpha_{i}K(\bs{x},\bs{\nu}_{i})$ where we must estimate the
means within $\frac{\epsilon}{2}$ error.

\noindent\textbf{Grid Search:}
Let us denote the parameters\footnote{To make our presentation simple we assume that the single parameter
variance is fixed and known. Note that it can also be estimated.} of the underlying mixture $p_o(\bs{x},\bs{\theta})$ by\\
 $\bs{\theta}=(\bs{m}, \bs{\alpha})=(\bs{\nu}_1,\ldots,\bs{\nu}_k,\bs{\alpha})\in \R^{k^2+k}$
and any approximating mixture $p_o(\bs{x},\tilde{\bs{\theta}})$ has parameters $\tilde{\bs{\theta}}=(\tilde{\bs{m}},\tilde{\bs{\alpha}})$.
We have proved the bounds  $f_1\left(d_{H}(\bs{m},\tilde{\bs{m}})\right)\leq\|p(\bs{x},\bs{\theta})-p(\bs{x},\tilde{\bs{\theta}})\|\leq f_2(d_{H}(\bs{m},
\tilde{\bs{m}}) + \|\bs{\alpha}  - \tilde{\bs{\alpha}}\|_1)$  (see Theorem \ref{thm:multivariate_lower_bound}, Lemma \ref{thm:multivariate_upper_bound}), where $f_1$ and $f_2$ are increasing functions.
Let $G$ be the step/grid size (whose value we need to set) that we use for gridding along each of the $k^2+k$
parameters over the grid $M_G$. We note that the $L^2$ norm of the difference can be computed efficiently by multidimensional trapezoidal rule or any other standard numerical analysis technique (see e.g., \cite{burden93}). Since this integration needs to be preformed on a $(k^2+k)$-dimensional space, for any pre-specified  precision parameter $\epsilon$, this can be done in time $\left(\frac{1}{\epsilon}\right)^{O(k^2)}$. 
Now note that there exists a point $\bs{\theta}^*=(\bs{m}^*,\bs{\alpha}^*)$ on the
grid $M_G$ , such that if somehow we can identify this point as our parameter estimate then  we make an error at
most $G/2$ in estimating each mixing weight and make an error at most $G\sqrt{k}/2$ in estimating each mean.
Since there are $k$ mixing weights and $k$ means to be estimated,
$\|p_o(\bs{x},\bs{\theta})-p_o(\bs{x},\bs{\theta}^*)\|\leq f_2(d_{H}(\bs{m},
\bs{m}^*) + \|\bs{\alpha}  - \bs{\alpha}^*\|_1)\leq f_2(G)=\frac{k\sqrt{1+k/\sigma^2}}{2(2\pi\sigma^2)^{k/2}}G$. Thus,
\[f_1\left(d_H(\bs{m},\bs{m}^*)\right)\leq\|p_o(\bs{x},\bs{\theta})-p_o(\bs{x},\bs{\theta}^*)\|\leq f_2(G)\]
Now, according to Lemma \ref{lem:kde_estimate}, using a sample of size $\Omega\left(\left[\frac{\log (2/\delta)}{\epsilon_*^2}\right]^k\right)$ we can obtain a kernel
density estimate such that with probability greater than $1-\frac{\delta}{2}$,
\begin{equation}\label{eq:kde_error}
\|p_{kde}-p_o(\bs{x},\bs{\theta})\|\leq \epsilon_*
\end{equation}
By triangular inequality this implies,
\begin{equation}\label{eq:proxy_error}
f_1\left(d_H(\bs{m},\bs{m}^*)\right)-\epsilon_*\leq\|p_{kde}-p_o(\bs{x},\bs{\theta}^*)\|\leq f_2(G)+\epsilon_*
\end{equation}
Since there is a one-to-one correspondence between the set of means of $\bs{m}$ and $\bs{m}^*$, $d_{H}(\bs{m},\bs{m}^*)$ essentially
provides the maximum estimation error for any pair of true mean and its corresponding estimate. Suppose we choose $G$ such that it satisfies
\begin{equation}\label{eq:choice_G}
2\epsilon_*+f_2(G)\leq f_1\left(\frac{\epsilon}{2}\right)
\end{equation}

For this choice of grid size, Equation \ref{eq:proxy_error} and Equation \ref{eq:choice_G} ensures that
$f_1\left(d_H(\bs{m},\bs{m}^*)\right)\leq f_2(G)+2\epsilon_*\leq f_1\left(\frac{\epsilon}{2}\right)$.
Hence $d_H(\bs{m},\bs{m}^*)\leq\frac{\epsilon}{2}$.
Now consider a point $\bs{\theta}^N=(\bs{m}^N,\bs{\alpha}^N)$ on the grid $M_G$ such that $d_H(\bs{m},\bs{m}^N)>\frac{\epsilon}{2}$.
This implies,
\begin{equation}\label{eq:non_optimal}
f_1\left(d_H(\bs{m},\bs{m}^N)\right)>f_1\left(\frac{\epsilon}{2}\right)
\end{equation}
Now,\\
$\|p_o(\bs{x},\bs{\theta}^N)-p_{kde}\|\stackrel{a}{\geq}\|p_o(\bs{x},\bs{\theta}^N)-p_o(\bs{x},\bs{\theta})\|-\|p_o(\bs{x},\bs{\theta})-p_{kde}\|$\\
$~~~~~~~~~~~~~~~~~~~~~~~~~~\stackrel{b}{\geq}f_1\left(d_H(\bs{m},\bs{m}^N)\right)-\epsilon_*$\\
$~~~~~~~~~~~~~~~~~~~~~~~~~~\stackrel{c}{>}f_1\left(\frac{\epsilon}{2}\right)-\epsilon_*$\\
$~~~~~~~~~~~~~~~~~~~~~~~~~~\stackrel{d}{\geq}f_2(G)+\epsilon_*$\\
$~~~~~~~~~~~~~~~~~~~~~~~~~~\stackrel{e}{\geq}\|p_o(\bs{x},\bs{\theta}^*)-p_{kde}\|$\\
where, inequality a follows from triangular inequality, inequality b follows from Equation \ref{eq:kde_error}, strict
inequality c follows from Equation \ref{eq:non_optimal}, inequality d follows from Equation \ref{eq:choice_G} and
finally inequality e follows from Equation \ref{eq:proxy_error}. Setting $\epsilon_*=\frac{1}{3}f_1\left(\frac{\epsilon}{2}\right)$,
Equation \ref{eq:choice_G} and the above strict inequality guarantees that for a choice of Grid size $G=f_2^{-1}\left(\frac{1}{3}f_1\left(\frac{\epsilon}{2}\right)\right)=\left(\frac{\alpha_{\min}^{4k}}{k^{3/2}}\right)\left(\frac{\epsilon}{8nk^2}\right)^{C_1k^2}$ the solution obtained by equation \ref{eq:mean_selection}
can have mean estimation error at most $\frac{\epsilon}{2}$.
Once projected onto SVD space each projected mean lies within a cube $[-\sqrt{\frac{n}{k}},\sqrt{\frac{n}{k}}]^k$.
With the above chosen grid size, grid search for the means runs in time $\left(\frac{k^{3/2}}{\alpha_{\min}^{4k}}\right)\cdot\left(\frac{n^{3/2}k^{1/2}}{\epsilon}\right)^{C_1k^2}$.
Note that grid search for the mixing weights runs in time
$\left(\frac{k^{3/2}}{\alpha_{\min}^{4k}}\right)\cdot\left(\frac{nk^2}{\epsilon}\right)^{C_1k^2}$.
\end{proof}

We now show that not only the mean estimates but also the mixing weights obtained by solving Equation \ref{eq:mean_selection}
satisfy $|\alpha_i-\tilde{\alpha}_i|\leq \epsilon$ for all $i$. In particular we show that if two mixtures have almost same means
and the $L^2$ norm of difference of their densities is small then the difference of the corresponding mixing weights
must also be small.

\begin{corollary}\label{cor:mixing_weight}
With sample size and grid size as in Theorem \ref{thm:main_theorem}, the solution of Equation \ref{eq:mean_selection} provides
mixing weight estimates which are, with high probability, within $\epsilon$ of their true values.
\end{corollary}
Due to space limitation we defer the proof to the Appendix.
\subsection{Lower Bound in 1-Dimensional Setting}
In this section we provide the proof of our main theoretical result in 1-dimensional setting.
Before we present the actual proof, we provide high level arguments that lead us to this result.
First note that  Fourier transform of a mixture of $k$ univariate Gaussians $q(x)=\sum_{i=1}^k\alpha_iK(x,\mu_i)$ is given by\\
$\mathcal{F}(q)(u)=\frac{1}{\sqrt{2\pi}}\int q(x)\exp(-iux)dx=\frac{1}{\sqrt{2\pi}}\sum_{j=1}^k\alpha_{j}\exp\left(-\frac{1}{2}(\sigma^2u^2+i2u\mu_j)\right)$\\
$~~~~~~~~~~~~=\frac{1}{\sqrt{2\pi}}\exp\left(-\frac{\sigma^2u^2}{2}\right)\sum_{j=1}^k\alpha_{j}\exp(-iu\mu_j)$\\
Thus, $\|\mathcal{F}(q)\|^2=\frac{1}{2\pi}\int|\sum_{j=1}^k\alpha_{j}\exp(-iu\mu_j)|^2\exp(-\sigma^2u^2)du$.
Since $L^2$ norm of a function and its Fourier transform are the same, we can write,\\
$\|q\|^2=\frac{1}{2\pi}\int{|\sum_{j=1}^k\alpha_{j}\exp(-iu\mu_j)|^2\exp(-\sigma^{2}u^{2})}du$.\\
Further,
$\frac{1}{2\pi}\int{|\sum_{j=1}^k\alpha_{j}\exp(-iu\mu_j)|^2\exp(-\sigma^{2}u^{2})}du=\frac{1}{2\pi}\int{|\sum_{j=1}^k\alpha_{j}\exp(iu\mu_j)|^2\exp(-\sigma^{2}u^{2})}du$
and we can write,
\[\|q\|^2=\frac{1}{2\pi}\int{|g(u)|^2\exp(-\sigma^{2}u^{2})}du\]
where $g(u)=\sum_{j=1}^{k}\alpha_{j}\exp(i\mu_{j}u)$. This a complex valued function of a real
variable which is infinitely differentiable everywhere. In order to bound the above square norm
from below, now our goal is to find an interval where $|g(u)|^{2}$ is bounded away from zero. In
order to achieve this, we write Taylor series expansion of $g(u)$ at the origin using $(k-1)$ terms.
This can be written in matrix vector multiplication format $g(u)=\boldsymbol{u^{t}A\alpha}+O(u^{k})$, where $\boldsymbol{u^{t}}=[1~u~\frac{u^{2}}{2!}\cdots~\frac{u^{k-1}}{(k-1)!}]$, such that $\boldsymbol{A\alpha}$
captures the function value and $(k-1)$ derivative values at origin.
In particular, $\|\bs{A}\bs{\alpha}\|^2$ is the sum of the squares of the function $g$ and $k-1$ derivatives at origin.
Noting that $\bs{A}$ is a Vandermonde
matrix we  establish (see Lemma \ref{lem:vandermonde_function_bound})
$\|\bs{A}\bs{\alpha}\|\geq\alpha_{\min}\left(\frac{t}{2\sqrt{n}}\right)^{k-1}$.
 This implies that at least one of the $(k-1)$ derivatives, say the $j^{th}$ one, of $g$ is bounded away
 from zero at origin. Once this fact is established, and noting that $(j+1)^{th}$ derivative of $g$ is
 bounded from above everywhere, it is easy to show (see Lemma~\ref{lem:derivative_bound_1}) that it
 is possible to find an interval $(0,a)$ where $j^{th}$ derivative of $g$ is bounded away from
 zero in this whole interval. Then using Lemma~\ref{lem:function_bound}, it can be shown that, it is
 possible to find a subinterval of $(0,a)$ where the $(j-1)^{th}$ derivative of $g$ is bounded
 away from zero. And thus, successively repeating this Lemma $j$ times, it is easy to show that there
 exists a subinterval of $(0,a)$ where $|g|$ is bounded away from zero. Once this subinterval is found,
 it is easy to show that $\|q\|^2$ is lower bounded as well.

Now we present the formal statement of our result.
\begin{theorem}[Lower bound in $\R$]\label{thm:1d_mixture_lower_bound}
Consider a mixture of $k$ univariate Gaussians $q(x)=\sum_{i=1}^{k}\alpha_{i}K(x,\mu_i)$
where, for all $i$, the mixing coefficients $\alpha_{i}\in (-1,1)$ and the means $\mu_{i}\in[-\sqrt{n},\sqrt{n}]$.
Suppose there exists a $\mu_l$ such that $\min_j|\mu_l-\mu_j|\geq t$, and for all $i, |\alpha_{i}|\geq\alpha_{\min}$. Then the $L^{2}$ norm of $q$ satisfies $||q||^2\geq\alpha_{\min}^{2k}\left(\frac{t}{n}\right)^{Ck^{2}}$ where $C$ is some positive constant independent of $k$.
\end{theorem}

\begin{proof}
Note that,
\[\|q\|^2=\frac{1}{2\pi}\int {|g(u)|^{2}\exp(-\sigma^{2}u^{2})du}\]
where, $g(u)=\sum_{j=1}^{k}\alpha_{j}\exp(i\mu_{j}u)$.
Thus, in order to bound the above square norm from below, we need to find an interval where $g(u)$ is bounded away from zero. Note that $g(u)$ is an infinitely differentiable function with $n^{th}$ order derivative
\footnote{Note that Fourier transform is closely related to the characteristics function and the $n^{th}$
derivative of $g$ at origin is  related to the $n^{th}$ order moment of the mixture in the Fourier domain.}
 $g^{(n)}(u)=\sum_{j=1}^{k}\alpha_{j}(i\mu_{j})^{n}\exp(i\mu_{j}u)$. Now we can write the Taylor series expansion of $g(u)$ about origin as, \[g(u)=g(0)+g^{(1)}(0)\frac{u}{1!}+g^{(2)}(0)\frac{u^{2}}{2!}+...+g^{(k-1)}(0)\frac{u^{(k-1)}}{(k-1)!}+O(u^{k})\]
which can be written as

\[g(u)=\left[\begin{array}{ccccc}
1 &u &\frac{u^{2}}{2!}&\cdots& \frac{u^{k-1}}{(k-1)!} \end{array}\right]
\underbrace{\left[\begin{array}{ccccc}
1 & 1 &1 & \cdots & 1\\
i\mu_{1}& i\mu_{2}& i\mu_{3} & \cdots & i\mu_{k}\\
(i\mu_{1})^{2}& (i\mu_{2})^{2}& (i\mu_{3})^{2} & \cdots & (i\mu_{k})^{2}\\
\cdots & \cdots & \cdots& \cdots & \cdots\\
\cdots & \cdots & \cdots& \cdots & \cdots\\
(i\mu_{1})^{k-1}& (i\mu_{2})^{k-1}& (i\mu_{3})^{k-1} & \cdots & (i\mu_{k})^{k-1} \end{array}\right]}_{\boldsymbol{A}}
\underbrace{\left[\begin{array}{c}
\alpha_{1}\\
\alpha_{2}\\
\cdot \\
\cdot\\
\alpha_{k} \end{array}\right]}_{\boldsymbol{\alpha}}
+O(u^{k})
\]
Note that matrix $\boldsymbol{A}$ is Vandermonde matrix thus, using Lemma~\ref{lem:vandermonde_function_bound} this implies $|g(0)|^{2}+|g^{(1)}(0)|^{2}+\cdots+|g^{(k-1)}(0)|^{2}=\|\boldsymbol{A\alpha}\|^{2}\geq \alpha_{\min}^{2}\left(\frac{t}{1+\sqrt{n}}\right)^{2(k-1)}\geq\alpha_{\min}^{2}\left(\frac{t}{2\sqrt{n}}\right)^{2(k-1)}$. This further implies that either $|g(0)|^{2}\geq\frac{\alpha_{\min}^{2}}{k}\left(\frac{t}{2\sqrt{n}}\right)^{2(k-1)}$ or there exists a $j\in\{1,2,\cdots,k-1\}$ such that $|g^{(j)}(0)|^{2}\geq\frac{\alpha_{\min}^{2}}{k}\left(\frac{t}{2\sqrt{n}}\right)^{2(k-1)}$. In the worst case we can have $j=k-1$, i.e. the $(k-1)$-th derivative of $g$ is lower bounded at origin and we need to find an interval where $g$ itself is lower bounded.

Next, note that for any $u, g^{(k)}(u)=\sum_{j=1}^{k}\alpha_{j}(i\mu_{j})^{k}\exp(iu\mu_{j})$. Thus, $|g^{(k)}|\leq \sum_{j=1}^{k}|\alpha_{j}||(i\mu_{j})^{k}|\leq \alpha_{\max}(\sqrt{n})^{k}$. Assuming $t\leq 2\sqrt{n}$, if we let $M=\frac{\alpha_{\min}}{\sqrt{k}}\left(\frac{t}{2\sqrt{n}}\right)^{k}$, then using Lemma~\ref{lem:derivative_bound_1}, if we choose $a=\frac{M}{2\sqrt{2}\alpha_{\max}(\sqrt{n})^{k}}=\frac{\alpha_{\min}}{\alpha_{\max}2\sqrt{2k}}\left(\frac{t}{2n}\right)^{k}$, and thus, in the interval $[0,a], |g^{(k-1)}|>\frac{M}{2}=\frac{\alpha_{\min}}{2\sqrt{k}}\left(\frac{t}{2\sqrt{n}}\right)^{k}$. This implies $|Re[g^{(k-1)}]|^{2}+|Im[g^{(k-1)}]|^{2}>\frac{\alpha_{\min}^{2}}{4k}\left(\frac{t}{2\sqrt{n}}\right)^{2k}$. For simplicity denote by $h=Re[g]$, thus, $h^{(k-1)}=Re[g^{(k-1)}]$ and without loss of generality assume $|h^{(k-1)}|>\frac{\alpha_{\min}}{2\sqrt{2k}}\left(\frac{t}{2\sqrt{n}}\right)^{k}=\frac{M}{2\sqrt{2}}$ in the interval $(0,a)$. Now repeatedly applying Lemma~\ref{lem:function_bound} $(k-1)$ times yields that in the interval $\left(\frac{(3^{k-1}-1)}{3^{k-1}}a,a\right)$,  (or in any other subinterval of length $\frac{a}{3^{k-1}}$ within $[0,a]$)\\ $|h|>\frac{M}{2\sqrt{2}}(\frac{a}{6})(\frac{a}{6.3})(\frac{a}{6.3^{2}})\cdots(\frac{a}{6.3^{k-1}})=\left(\frac{M}{2\sqrt{2}}\right)\left(\frac{a}{6}\right)^{k}\left(\frac{1}{3^{\frac{k(k-1)}{2}}}\right)=\frac{\alpha_{\max}(\sqrt{n})^{k}a^{k+1}}{2^{k}3^{\frac{k^{2}+k}{2}}}$\\
In particular, this implies, $|g|^{2}\geq|h|^{2}>\frac{\alpha_{\max}^{2}n^{k}a^{2(k+1)}}{2^{2k}3^{k^{2}+k}}$ in an interval $\left(\frac{(3^{k-1}-1)}{3^{k-1}}a,a\right)$.

 Next, note that $0<a\leq 1 \Rightarrow  \exp(-\sigma^{2})\leq\exp(-\sigma^{2}a^{2})$. Now, denoting $\beta_{1}=\frac{(3^{k-1}-1)}{3^{k-1}}a, \beta_{2}=a$, we have,\\
$\|q\|^2\geq\frac{1}{2\pi}\int_{\beta_{1}}^{\beta_{2}}{|g(u)|^{2}\exp(-\sigma^{2}u^{2})du}\geq\frac{\beta_{2}-\beta_{1}}{2\pi}|g(\beta_{2})|^{2}\exp(-\sigma^{2})$\\
$=\left(\frac{\exp(-\sigma^{2})}{2\pi}\right)\frac{\alpha_{\max}^{2}n^{k}a^{2k+3}}{2^{2k}3^{k^{2}+2k-1}}=\left(\frac{\exp(-\sigma^{2})\alpha_{\min}^{2k+3}}{2\pi}\right)\left(\frac{t^{2k^{2}+3k}}{2^{2k^{2}+5k+9/2}3^{k^{2}+2k-1}(\alpha_{\max})^{2k+1}k^{k+3/2}n^{2k^{2}+2k}}\right)$\\
$\geq\left(\frac{\exp(-\sigma^{2})\alpha_{\min}^{2k+3}}{2\pi}\right)\left(\frac{t^{2k^{2}+3k}}{2^{2k^{2}+5k+9/2}3^{k^{2}+2k-1}k^{k+3/2}n^{2k^{2}+2k}}\right)$\\
$\geq \alpha_{\min}^{2k}\left(\frac{t^{2k^{2}+3k}}{2^{O(k^{2}\log n)}}\right)=\alpha_{\min}^{2k}\left(\frac{t}{n}\right)^{O(k^{2})}$\\
where the last inequality follows from the fact that if we let,\\ $F(k)=2^{2k^{2}+5k+9/2}3^{k^{2}+2k-1}k^{k+3/2}n^{k^{2}+2k}$ then taking $\log$ with base 2 on both sides yields,\\
$\log(F(k))=(2k^{2}+5k+9/2)+(k^{2}+2k-1)\log 3+\left(k+3/2\right)\log k+(2k^{2}+2k)\log n=O(k^{2}\log n)$.\\
Thus, $F(k)=2^{O(k^{2}\log n)}=n^{O(k^{2})}$.
\end{proof}

\subsection{Determinant of Vandermonde Like Matrices}
In this section we derive a result for the determinant of a Vandermonde-like matrix.
This result will be useful in finding the angle made by any column of a Vandermonde matrix to the space spanned by the rest of the
columns and will be useful in deriving the lower bound in Theorem \ref{thm:1d_mixture_lower_bound}.

Consider any $(n+1)\times n$ matrix $B$ of the form
\[B=
\left[\begin{array}{ccccc}
1 & 1 &1 & \cdots & 1\\
x_{1}& x_{2}& x_{3} & \cdots & x_{n}\\
x_{1}^{2}& x_{2}^{2}& x_{3}^{2} & \cdots & x_{n}^{2}\\
\cdots & \cdots & \cdots& \cdots & \cdots\\
\cdots & \cdots & \cdots& \cdots & \cdots\\
x_{1}^{n}& x_{2}^{n}& x_{3}^{n} & \cdots & x_{n}^{n} \end{array}\right]
\]
If the last row is removed then it exactly becomes an $n\times n$ Vandermonde matrix having determinant $\Pi_{i>j}(x_{i}-x_{j})$. The interesting fact is that if any other row except the last one is removed then the corresponding $n\times n$ matrix has a structure very similar to that of a Vandermonde matrix. The following result shows how the determinants of such matrices are related to $\Pi_{i>j}(x_{i}-x_{j})$.
\begin{lemma}\label{lem:vandermonde_like}
For $1\leq i\leq (n-1)$, let $B_{i}$ represents the $n\times n$ matrix obtained by removing the $i^{th}$ row from $B$. Then $\det(B_{i})=c_{i}\Pi_{s>t}(x_{s}-x_{t})$ where $c_{i}$ is a polynomial having $n \choose i-1$ terms with each term having degree $(n-i+1)$. Terms of the polynomial $c_i$ represent the possible ways in which $(n-i+1)$ $x_{j}$s can be chosen from $\{x_{i}\}_{i=1}^{n}$.
\end{lemma}
\begin{proof}
First note that if a matrix has elements that are monomials in some set of variables, then its determinant will in general be polynomial in those variables. Next, by the basic property of a determinant, that it is zero if two of its columns are same, we can deduce that for $1\leq i<n, \det(B_{i})=0$ if $x_{s}=x_{t}$ for some $s\neq t, 1\leq s,t<n$,  and hence $q_{i}(x_{1},x_{2},...,x_{n})=\det(B_{i})$ contains a factor $p(x_{1}, x_{2},...,x_{n})=\Pi_{s>t}(x_{s}-x_{t})$. Let $q_{i}(x_{1},x_{2},...,x_{n})=p(x_{1},x_{2},...,x_{n})r_{i}(x_{1},x_{2},...,x_{n})$.

Now, note that each term of $p(x_{1}, x_{2},...,x_{n})$ has degree $0+1+2+...+(n-1)=\frac{n(n-1)}{2}$. Similarly, each term of the polynomial $q_{i}(x_{1},x_{2},...,x_{n})$ has degree $(0+1+2+...+n)-(i-1)=\frac{n(n+1)}{2}-(i-1)$. Hence each term of the polynomial $r_{i}(x_{1},x_{2},...,x_{n})$ must be of degree $\frac{n(n+1)}{2}-(i-1)-\frac{n(n-1)}{2}=(n-i+1)$. However in each term of $r_{i}(x_{1},x_{2},...,x_{n})$, the maximum power of any $x_{j}$ can not be greater than 1. This follows from the fact that maximum power of $x_{j}$ in any term of $q_{i}(x_{1},x_{2},...,x_{n})$ is $n$ and in any term of $p(x_{1},x_{2},...,x_{n})$ is $(n-1)$. Hence each term of $r_{i}(x_{1},x_{2},...,x_{n})$ consists of $(n-i+1)$ different $x_{j}$s and represents the different ways in which $(n-i+1)$ $x_{j}$s can be chosen from $\{x_{i}\}_{i=1}^{n}$. And since it can be done in $n \choose n-i+1$=$n \choose i-1$ ways there will be $n \choose i-1$ terms in $r_{i}(x_{1},x_{2},...,x_{n})$.
\end{proof}

\subsection{Estimation of Unknown Variance}\label{sec:variance_estimation}
In this section we discuss a procedure  for consistent estimation of the
unknown variance due to~\cite{lindsay89} (for the one-dimensional case) and will prove that the estimate is 
polynomial. This estimated variance can then be used
in place of true variance in our main algorithm discussed earlier and the remaining mixture parameters can be estimated subsequently. 

We start by noting   a mixture of $k$ identical spherical Gaussians $\sum_{i=1}^k\alpha_i\mathcal{N}(\bs{\mu}_i, \sigma^2I)$ in $\R^n$
 projected on an arbitrary  line becomes a mixture of identical 1-dimensional Gaussians $p(x)=\sum_{i=1}^k\alpha_i\mathcal{N}(\mu_i, \sigma^2)$.
While the means of components may no longer be different, the variance does not change. Thus, the problem is easily reduced to the 1-dimensional case.

%
%
%
We will now show that  the variance of a mixture of $k$ Gaussians in $1$ dimension
can be estimated from a sample
 of size $\poly\left(\frac{1}{\epsilon},\frac{1}{\delta}\right)$, where $\epsilon>0$ is the precision ,with probability $1-\delta$ in time $\poly\left(\frac{1}{\epsilon},\frac{1}{\delta}\right)$. 
This will lead to an estimate for the $n$-dimensional mixture using $\poly\left(n,\frac{1}{\epsilon},\frac{1}{\delta}\right)$ sample points/operations.




Consider now the set of Hermite polynomials  $\gamma_i(x,\tau)$
given by the recurrence relation $\gamma_i(x,\tau)=x\gamma_{i-1}(x,\tau)-(i-1)\tau^2\gamma_{i-2}(x,\tau)$, 
where $\gamma_0(x,\tau)=1$ and $\gamma_1(x,\tau)=x$. Take  $M$ to be the $(k+1)\times (k+1)$ matrix defined by 
$$
M_{ij} = \mathbb{E}_{p}[\gamma_{i+j}(X,\tau)], ~~ 0\le i+j \le 2k.
$$ 
It is shown in Lemma 5A of~\cite{lindsay89} that the determinant $\det(M)$ is a polynomial in $\tau$ and, moreover, that the smallest 
positive root of $\det(M)$, viewed is a function of $\tau$, is equal to the variance $\sigma$ of the original mixture $p$. 
We will use $d(\tau)$ to represent $\det(M)$.

This result leads to an estimation 
procedure, after observing that $\mathbb{E}_{p}[\gamma_{i+j}(X,\tau)]$ can be replaced by its empirical value given a sample $X_1, X_2,...,X_N$ from 
the  mixture distribution $p$. Indeed, one can construct the empirical version of the matrix $M$ by putting 
\begin{equation}\label{eq_emp_moments}
\hat{M}_{ij} = \frac{1}{N}\sum_{t=1}^N [\gamma_{i+j}(X_{t},\tau)], ~~ 0\le i+j \le 2k.
\end{equation}
It is clear that $\hat{d}(\tau) = \det(\hat{M})(\tau)$ is a polynomial in $\tau$. Thus we can provide an estimate ${\sigma}^*$ 
for the variance $\sigma$ by taking the smallest positive root of $\hat{d}(\tau)$. This leads to the following estimation procedure :\\

\noindent\textbf{Parameter:} Number of components  $k$.\\
\textbf{Input:} $N$ points in $\mathbb{R}^n$ sampled from $\sum_{i=1}^k\alpha_i\mathcal{N}(\bs{\mu}_i, \sigma^2I)$.\\
\textbf{Output:} $\sigma^*$, estimate of the unknown variance.\\

\textbf{Step 1.} Select an arbitrary direction $\bs{v}\in\mathbb{R}^n$ and project the data points onto this direction.

\textbf{Step 2.} Construct the $(k+1)\times(k+1)$ matrix $\hat{M}(\tau)$ using Eq.~\ref{eq_emp_moments}

\textbf{Step 3.} Compute the polynomial  $\hat{d}(\tau) = \det(\hat{M})(\tau)$. Obtain the estimated variance $\sigma^*$ by approximating the 
smallest positive root of $\hat{d}(\tau)$. This can be done efficiently by using any standard numerical method or even a grid search.

We will now state our main result in this section, which establishes that this algorithm for variance estimation is indeed polynomial in both the 
ambient dimension $n$ and the  inverse of the   desired accuracy $\epsilon$.

\begin{theorem}\label{thm:variance_estimation}
For any $\epsilon>0, 0<\delta<1$, if sample size $N>O\left(\frac{n^{\poly(k)}}{\epsilon^2\delta}\right)$,
then the above procedure 
provides an
estimate $\sigma^*$ of the unknown variance $\sigma$ such that  $|\sigma-\sigma^*|\leq\epsilon$ with probability greater than $1-\delta$.
\end{theorem}

The idea of the proof is to show that the coefficients of the polynomials ${d}(\tau)$ and $\hat{d}(\tau)$ are polynomially close, given enough samples from $p$.
That (under some additional technical conditions) can be shown to imply that the smallest positive roots of these polynomials are also close.
To verify that ${d}(\tau)$ and $\hat{d}(\tau)$ are close, we use the fact that the coefficients of ${d}(\tau)$ are polynomial functions of the first $2k$ moments of $p$,
while coefficients of $\hat{d}(\tau)$ are the same functions of the empirical moment estimates. Using standard concentration inequalities for the first 
$2k$ moments and providing a bound for these functions the result.

%
%
%
%

The details of the proof are provided in the Appendix \ref{app:variance}.

\bibliographystyle{plain}
\bibliography{current_work}
\newpage
\appendix
\section*{Appendix}
\section{Proof of Some Auxiliary Lemmas}
\begin{lemma}\label{lem:angle_bound}
For any $\bs{v}_{1}, \bs{v}_{2} \in \mathbb{R}^{n}$ and any $\alpha_{1}, \alpha_{2} \in \mathbb{R}, \|\alpha_{1}\bs{v}_{1}+\alpha_{2}\bs{v}_{2}\|\geq |\alpha_{1}||\bs{v}_{1}||\sin(\beta)|$ where $\beta$ is the angle between $\bs{v}_{1}$ and $\bs{v}_{2}$.
\end{lemma}
\begin{proof}
Let $s\in\mathbb{R}$ such that $\langle\alpha_{1}\bs{v}_{1}+s\bs{v}_{2}, \bs{v}_{2}\rangle=0$. This implies $s=-\frac{\alpha_{1}\langle\bs{v}_{1}, \bs{v}_{2}\rangle}{\|\bs{v}_{2}\|^{2}}$.Now,\\
$\|\alpha_{1}\bs{v}_{1}+\alpha_{2}\bs{v}_{2}\|^{2}=\|(\alpha_{1}\bs{v}_{1}+s\bs{v}_{2})+(\alpha_{2}-s)\bs{v}_{2}\|^{2}=\|\alpha_{1}\bs{v}_{1}+s\bs{v}_{2}\|^{2}+\|(\alpha_{2}-s)\bs{v}_{2}\|^{2}\geq\|\alpha_{1}\bs{v}_{1}+s\bs{v}_{2}\|^{2}$\\
$~~~~~~~~~~~~~~~~~~~~~=\langle\alpha_{1}\bs{v}_{1}+s\bs{v}_{2},\alpha_{1}\bs{v}_{1}+s\bs{v}_{2}\rangle=\langle\alpha_{1}\bs{v}_{1}+s\bs{v}_{2},\alpha_{1}\bs{v}_{1}\rangle=\alpha_{1}^{2}\|\bs{v}_{1}\|^{2}+\alpha_{1}s\langle\bs{v}_{1}, \bs{v}_{2}\rangle$\\
$~~~~~~~~~~~~~~~~~~~~~=\alpha_{1}^{2}\|\bs{v}_{1}\|^{2}-\alpha_{1}\left(\frac{\alpha_{1}\langle\bs{v}_{1}, \bs{v}_{2}\rangle}{\|\bs{v}_{2}\|^{2}}\right)\langle\bs{v}_{1}, \bs{v}_{2}\rangle=\alpha_{1}^{2}\|\bs{v}_{1}\|^{2}-\frac{\alpha_{1}^{2}}{\|\bs{v}_{2}\|^{2}}\left(\|\bs{v}_{1}\|~\|\bs{v}_{2}\|\cos(\beta)\right)^{2}$\\
$~~~~~~~~~~~~~~~~~~~~~=\alpha_{1}^{2}\|\bs{v}_{1}\|^{2}(1-\cos^{2}(\beta))=\alpha_{1}^{2}\|\bs{v}_{1}\|^{2}\sin^{2}(\beta)$
\end{proof}
\begin{lemma}\label{lem:derivative_bound_1}
Let $h: \mathbb{R}\rightarrow\mathbb{C}$ be an infinitely differentiable function such that for some positive integer $n$ and real $M,T>0, |h^{(n)}(0)|>M$ and $|h^{(n+1)}|<T$. Then for any $0<a<\frac{M}{T\sqrt{2}}$, $|h^{(n)}|>M-\sqrt{2}Ta$ in the interval $[0,a]$.
\end{lemma}
\begin{proof}
Using mean value theorem for complex valued function, for any $x\in[0,a], |h^{(n)}(x)-h^{(n)}(0)|\leq \sqrt{2}Ta$, which implies $M-|h^{(n)}(x)|<\sqrt{2}Ta$.
\end{proof}

\begin{lemma}\label{lem:function_bound}
Let $h: \mathbb{R}\rightarrow\mathbb{R}$ be an infinitely differentiable function such that for some positive integer $n$ and real $M>0, |h^{(n)}|>M$ in an interval $(a,b)$. Then $|h^{(n-1)}|>M(b-a)/6$ in a smaller interval either in $(a,\frac{2a+b}{3})$ or in $(\frac{a+2b}{3},b)$.
\end{lemma}
\begin{proof}
Consider two intervals $I_{1}=(a,\frac{2a+b}{3})$ and $I_{2}=(\frac{a+2b}{3},b)$. Chose any two arbitrary points $x\in I_{1}, y\in I_{2}$. Then by mean value theorem, for some $c \in (a,b), |h^{(n-1)}(x)-h^{(n-1)}(y)|=|h^{(n)}(c)||x-y|>M(b-a)/3$.

If the statement of the Lemma is false then we can find $x_{*} \in I_{1}$ and $y_{*} \in I_{2}$ such that $|h^{(n-1)}(x_{*})|\leq M(b-a)/6$ and $|h^{(n-1)}(y_{*})|\leq M(b-a)/6$. This implies $|h^{(n-1)}(x_{*})-h^{(n-1)}(y_{*})|\leq M(b-a)/3$. Contradiction.
\end{proof}

\noindent\textbf {Generalized cross product:}\\
Cross product between two vectors $\bs{v}_{1}, \bs{v}_{2}$  in $\mathbb{R}^{3}$ is a vector orthogonal to the space spanned by $\bs{v}_{1}, \bs{v}_{2}$. This idea can be generalized to any finite dimension in terms of determinant and inner product as follows. The cross product of $(n-1)$ vectors $\bs{v}_{1},...,\bs{v}_{n-1}\in\mathbb{R}^{n}$ is the unique vector $\bs{u}\in\mathbb{R}^{n}$ such that for all $\bs{z}\in\mathbb{R}^{n}, \langle\bs{z}, \bs{u}\rangle=\det[\bs{v}_{1},...,\bs{v}_{n-1},\bs{z}]$. With this background we provide the next result for which we introduce the following $k\times k$ Vandermonde matrix $A$.
\[A=
\left[\begin{array}{ccccc}
1 & 1 &1 & \cdots & 1\\
x_{1}& x_{2}& x_{3} & \cdots & x_{k}\\
x_{1}^{2}& x_{2}^{2}& x_{3}^{2} & \cdots & x_{k}^{2}\\
\cdots & \cdots & \cdots& \cdots & \cdots\\
\cdots & \cdots & \cdots& \cdots & \cdots\\
x_{1}^{k-1}& x_{2}^{k-1}& x_{3}^{k-1} & \cdots & x_{k}^{k-1} \end{array}\right]
\]
\begin{lemma}\label{lem:vandermonde_function_bound}
For any integer $k>1$, and positive $a,t\in\R$, let $x_{1}, x_{2},...,x_{k}\in[-a,a]$  and there exists an $x_{i}$ such
that  $t=\min_{j,j\neq i}|x_{i}-x_{j}|$. Let $\bs{\alpha}=(\alpha_{1},\alpha_{2},...,\alpha_{k})\in\mathbb{R}^{k}$ with $\min_{i}|\alpha_{i}|\geq\alpha_{\min}$. Then for $A$ as defined above, $\|A\bs{\alpha}\|\geq\alpha_{\min}\left(\frac{t}{1+a}\right)^{k-1}$.
\end{lemma}
\begin{proof}
We will represent the $i^{th}$ column of $A$ by $\bs{v}_{i}\in\mathbb{R}^{k}$. Without loss of generality, let the nearest point to $x_{k}$ be at a distance $t$. Then $\|A\bs{\alpha}\|=\|\alpha_{k}\bs{v}_{k}+\sum_{i=1}^{k-1}\alpha_{i}\bs{v}_{i}\|$. Note that $\sum_{i=1}^{k-1}\alpha_{i}\bs{v}_{i}$ lies in the space spanned by the vectors $\{\bs{v}_{i}\}_{i=1}^{k-1}$, i.e., in $span\{\bs{v}_{1}, \bs{v}_{2},..., \bs{v}_{k-1}\}$. Let $\bs{u}\in\mathbb{R}^{k}$ be the vector orthogonal to $span\{\bs{v}_{1}, \bs{v}_{2},..., \bs{v}_{k-1}\}$ and represents the cross product of $\bs{v}_{1}, \bs{v}_{2},...,\bs{v}_{k-1}$. Let $\beta$ be the angle between $\bs{u}$ and $\bs{v}_{k}$. Then using Lemma \ref{lem:angle_bound},  $\|A\bs{\alpha}\|\geq |\alpha_{k}|~\|\bs{v}_{k}\|~|\sin(90-\beta)|\geq\alpha_{\min}\|\bs{v}_{k}\|~|\cos(\beta)|$.
Using the concept of generalized cross product
\begin{equation}\label{eq:cross_product_norm}
\langle\bs{u}, \bs{v}_{k}\rangle=\det(A)\Rightarrow \|\bs{v}_{k}\|~|\cos(\beta)|=\frac{|\det(A)|}{\|\bs{u}\|}
\end{equation}
Let $\tilde{A}=[\bs{v}_{1}, \bs{v}_{2}, ..., \bs{v}_{k-1}]\in\R^{k\times(k-1)}$. Note that $\|u\|^{2}=\sum_{i=1}^{k}(\det(\tilde{A}_{i}))^{2}$ where $\tilde{A}_{i}$ represents the $(k-1)\times(k-1)$ matrix obtained by removing the $i^{th}$ row from $\tilde{A}$. Since each $|x_{i}|\leq a$, and  for any integer $0\leq b\leq k-1, {k-1 \choose b}={k-1 \choose k-1-b}$, using Lemma \ref{lem:vandermonde_like},\\
$\|\bs{u}\|^{2}=\Pi_{k-1\geq s>t\geq 1}|x_{s}-x_{t}|^{2}\left(1 + \left({k-1 \choose 1}a\right)^{2} + \left({k-1 \choose 2}a^{2}\right)^{2} + ... + \left({k-1 \choose k-1}a^{k-1}\right)^{2}\right)$\\
$\leq\Pi_{k-1\geq s>t\geq 1}|x_{s}-x_{t}|^{2}\left(1 + {k-1 \choose 1}a + {k-1 \choose 2}a^{2}+ ... + {k-1 \choose k-1}a^{k-1}\right)^2$\\
$\leq \left(\Pi_{k-1\geq s>t\geq 1}|x_{s}-x_{t}|^{2}\right)(1+a)^{2(k-1)}$. \\
where, the first inequality follows from the fact that for any $b_1,b_2,\ldots,b_n>0, \sum_{i=1}^nb_i^2\leq\left(\sum_{i=1}^n b_i\right)^2$
and the second inequality follows from the fact that for any $c>0$, and positive integer $n, (c+1)^n=\sum_{i=0}^n{n \choose i}c^i$.
Since $\det(A)=\Pi_{k\geq s>t\geq 1}=\Pi_{k-1\geq s>t\geq 1}(x_{s}-x_{t})\Pi_{k-1\geq r\geq 1}(x_{k}-x_{r})$. Plugging these values in Equation \ref{eq:cross_product_norm} yields, $\|\bs{v}_{k}\|~|\cos(\beta)|\geq\frac{\Pi_{k-1\geq r\geq 1}(x_{k}-x_{r})}{(2a)^{k-1}}\geq\frac{t^{k-1}}{(1+a)^{k-1}}$. This implies, $\|A\bs{\alpha}\|\geq\alpha_{\min}\left(\frac{t}{1+a}\right)^{k-1}$.
\end{proof}

\begin{figure}[t]
\centerline{\includegraphics[scale=0.35]{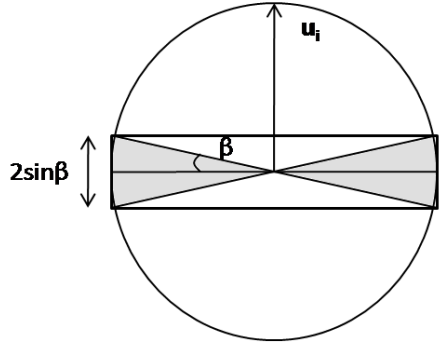}}
\caption{For $\bs{u}_i$, the shaded region represents the volume $\mathcal{C}_i$, which corresponds to all the ``bad" directions associated with $\bs{u}_i$. An upper bound for $\mathcal{C}_i$ is $\mathcal{C}_i^{'}$ which is a cylinder with radius 1 and height $2\sin\beta$ and is shown by the rectangular box.}
\label{fig:existence_of_direction}
\end{figure}

\begin{lemma}\label{lem:existence_of_direction}
Consider any set of $k$ points $\{\boldsymbol{x}_{i}\}_{i=1}^{k}$ in $\mathbb{R}^{n}$. There exists a direction $\bs{v}\in \mathbb{R}^{n}, \|\bs{v}\|=1$ such for any $i,j~~|\langle\bs{x}_{i}, \bs{v}\rangle-\langle\bs{x}_{j}, \bs{v}\rangle|>\frac{\|\bs{x}_{i}-\bs{x}_{j}\|}{k^{2}}$.
\end{lemma}
\begin{proof}
For $k$ points $\{\bs{x}_{i}\}_{i=1}^{k}$, there exists $k \choose 2$ directions $\frac{(\bs{x}_{i}-\bs{x}_{j})}{\|\bs{x}_{i}-\bs{x}_{j}\|}, i,j=1,2,\ldots,k$ obtained by joining all possible pair of points. Let us renumber these directions as $\bs{u}_{i}, i=1,2,...,$$k\choose 2$. Now, consider any arbitrary direction $\bs{u}_{j}$ formed using points $\boldsymbol{x}_{m}$ and $\boldsymbol{x}_{n}$ respectively. If $\bs{x}_{i}$s are projected to any direction orthogonal to $\boldsymbol{u}_{j}$, then at least two $\bs{x}_{i}$s, $\boldsymbol{x}_{m}$ and $\boldsymbol{x}_{n}$ coincide. In order to show that there exists some direction, upon projecting the $\bs{x}_{i}$s on which, no two $\bs{x}_{i}$s become too close, we adopt the following strategy. Consider a $n$ dimensional unit ball $\mathcal{S}$ centered at origin and place all $\boldsymbol{u}_{j}, j=1,2,\ldots,$$k\choose 2$ directions vectors on the ball starting at origin. Thus each $\bs{u}_{j}$ is represented by a $n$ dimensional point on the surface of $\mathcal{S}$.
For any $\boldsymbol{u}_{j}$, consider all vectors $v\in \mathbb{R}^{n}$ lying on a manifold,- the $(n-1)$ dimensional unit ball having center at origin and orthogonal to $\boldsymbol{u}_{j}$. These directions are the ``bad" directions because if $\bs{x}_{i}$s are projected to any of these directions then at least two $\bs{x}_{i}$s coincide. We want to perturb these ``bad" directions a little bit and form an object and show that we can control the size of an angle such that volume of union of these $k \choose 2$ objects is much less than the volume of $\mathcal{S}$, which implies that there are some ``good" directions for projection.

Consider any $0<\beta\leq \frac{\pi}{2}$. For any $\boldsymbol{u}_{i}$, let $\mathcal{C}_{i}=\{x\in\mathcal{S}: \arcsin\frac{|\langle \boldsymbol{x}, \boldsymbol{u}_{i}\rangle|}{\|\boldsymbol{x}\|}\leq \beta\}$. $\mathcal{C}_{i}$ is the perturbed version of a bad direction and we do not want to project $\bs{x}_{i}$s on any direction contained in $\mathcal{C}_{i}$. The volume of $\mathcal{C}_{i}$ is shown in the shaded area in Figure \ref{fig:existence_of_direction}. A simple upper bound of this volume can be estimated by the volume of a larger  $n$ dimensional cylinder $\mathcal{C}^{'}$ of radius $1$ and height $2\sin(\beta)$. Let $\mathcal{C}=\cup_{i=1}^{k \choose 2}\mathcal{C}_{i}$ . Thus, total volume of $\mathcal{C}$  is $vol(\mathcal{C})=\cup_{i=1}^{k \choose 2}vol(\mathcal{C}_{i})\leq\cup_{i=1}^{k \choose 2}vol(\mathcal{C}^{'}_{i})\leq k(k-1)\sin(\beta)\times\left(\frac{(\pi)^{\frac{(n-1)}{2}}}{\Gamma(\frac{n-1}{2}+1)}\right)$. Note that $vol(\mathcal{S})=\frac{(\pi)^{\frac{n}{2}}}{\Gamma(\frac{n}{2}+1)}$. We want $vol(\mathcal{C})<vol(\mathcal{S})$. This implies,
\begin{equation}\label{eq:vol_inequality}
\frac{k(k-1)\sin(\beta)}{\sqrt{\pi}}<\frac{\Gamma(\frac{n-1}{2}+1)}{\Gamma(\frac{n}{2}+1)}
\end{equation}
Now we consider two cases.
\\
\underline{case 1: $n$ is even} \\
From the definition of Gamma function denominator of r.h.s of Equation~\ref{eq:vol_inequality} is $\left(\frac{n}{2}\right)!$ Since $n-1$ is odd, using the definition of Gamma function, the numerator of Equation~\ref{eq:vol_inequality} becomes $\frac{\sqrt{\pi}(n-1)!!}{2^{\frac{n}{2}}}=\frac{2\sqrt{\pi}(n-1)!}{2^{n}(\frac{n-2}{2})!}$ using the fact that $(2n+1)!!=\frac{(2n+1)!}{2^{n}n!}$. Thus r.h.s of Equation~\ref{eq:vol_inequality} becomes $2\sqrt{\pi}\left(\frac{(n-1)!}{2^{n}(\frac{n}{2}-1)!(\frac{n}{2})!}\right)<2\sqrt{\pi}\times \frac{1}{2}=\sqrt{\pi}$, where the last inequality can be easily shown as follows,\\ $\left(\frac{(n-1)!}{2^{n}(\frac{n}{2}-1)!(\frac{n}{2})!}\right)=\frac{1}{2}\frac{(n-1)(n-2)(n-3)(n-4)\cdots 1}{[(n-2)(n-4)(n-6)\cdots 2][(n(n-2)(n-4)\cdots 2]}=\frac{1}{2}\left(\frac{(n-1)(n-3)(n-5)\cdots 1}{n(n-2)(n-4)\cdots 2}\right)< \frac{1}{2}$.\\
\\
\underline{case 2: $n$ is odd}\\
$n-1$ is even and thus numerator of r.h.s of Equation~\ref{eq:vol_inequality} becomes $\left(\frac{n-1}{2}\right)!$ The denominator become $\frac{\sqrt{\pi}n!!}{2^{\frac{n+1}{2}}}$, which in turn is equal to $\frac{\sqrt{\pi}n!}{2^{n}(\frac{n-1}{2})!}$ using the relation between double factorial and factorial. Thus, r.h.s of Equation~\ref{eq:vol_inequality} becomes $\frac{1}{\sqrt{\pi}}\left(\frac{2^{n}(\frac{n-1}{2})!(\frac{n-1}{2})!}{n!}\right)< \frac{1}{\sqrt{\pi}}\times 2=\frac{2}{\sqrt{\pi}}$. The last inequality follows from the fact that,\\
 $\left(\frac{2^{n}(\frac{n-1}{2})!(\frac{n-1}{2})!}{n!}\right)=\frac{2[(n-1)(n-3)(n-5)\cdots 2][(n-1)(n-3)(n-5)\cdots 2]}{n(n-1)(n-2)(n-3)(n-4)\cdots 1}=2\left(\frac{(n-1)(n-3)(n-5)\cdots 2}{n(n-2)(n-4)\cdots 3.1}\right)< 2$.

Thus, for any $n$, to ensure existence of a good direction, we must have $\frac{k(k-1)\sin(\beta)}{\sqrt{\pi}}<\sqrt{\pi}$ which implies $\sin(\beta)<\frac{\pi}{k(k-1)}$. Fixing $\beta$ small enough, in particular setting $\beta=\beta^*$ such that $\sin(\beta^*)=\frac{1}{k^{2}}$ satisfies strict inequality $\sin(\beta^*)<\frac{\pi}{k(k-1)}$.
Once $\beta$ is chosen this way, volume of $\mathcal{C}$ is less than volume of $\mathcal{S}$ and hence there exists some ``good" direction $\bs{v}, \|\bs{v}\|=1$, such that if $\bs{x}_{i}$s are projected along this ``good" direction $\bs{v}$, no two $\langle\bs{x}_{i},\bs{v}\rangle$ becomes too close.
Now consider any $\bs{v}$ on the surface of $\mathcal{S}$ which is not contained in any of the $\mathcal{C}^{'}_{i}$s and hence in any of the $\mathcal{C}_{i}$s. This implies for any $i,j, |\langle\bs{v},\frac{(\bs{x}_{i}-\bs{x}_{j})}{\|\bs{x}_{i}-\bs{x}_{j}\|}\rangle|> sin(\beta^*)=\frac{1}{k^{2}}$, and hence $|\langle\bs{v},(\bs{x}_{i}-\bs{x}_{j})\rangle|>\frac{\|\bs{x}_{i}-\bs{x}_{j}\|}{k^{2}}$.
\end{proof}

Note that the above Lemma can also be considered as a special kind of one sided version of Johnson-Lindenstraus Lemma, specifically, when equivalently expressed as,- for given small enough $\beta>0$ (hence $\sin(\beta)\approx\beta$), and vector $\bs{y}=\bs{x}_i-\bs{x}_j$, with probability at least $1-O(\beta)$, a random unit vector $\bs{v}$ has the property that the projection of $\bs{y}$ on to the span of $\bs{v}$ has length at least $\beta\|\bs{y}\|$. However, our result is deterministic.

\begin{lemma}\label{lem:norm_projection_bound}
Let $g:\mathbb{R}^{k}\rightarrow \mathbb{R}$ be a continuous bounded function. Let $\bs{v}, \bs{u}_{1},...,\bs{u}_{k-1}\in \mathbb{R}^{k}$ be an orthonormal basis of $\mathbb{R}^{k}$ and let $g_{1}:\mathbb{R}\rightarrow \mathbb{R}$ be defined as $g_{1}(v)=\int\cdots\int g(v,u_{1},...u_{k-1})du_{1}\cdots du_{k-1}$. Then for some $c>0, \|g\|^{2}\geq\left(\frac{1}{c\sigma}\right)^{k}\|g_{1}\|^{2}$.
\end{lemma}
\begin{proof}
Note that $\|g\|^{2}=\int\left(\int\cdots\int|g(v,u_{1},\cdots,u_{k-1})|^{2}du_{1}\cdots du_{k-1}\right)dv$ and,\\
$\|g_{1}\|^{2}=\int|g_{1}(v)|^{2}dv=\int\left(\int\cdots\int g(v,u_{1},\cdots,u_{k-1})du_{1},\cdots du_{k-1}\right)^{2}dv$.
For any sufficiently large $L>0$  we concentrate on a bounded domain $\mathcal{A}=[-L,L]^{k}\subset\mathbb{R}^{k}$ outside which the function value becomes arbitrarily small and so do the norms. Note that this is a very realistic assumption because component Gaussians have exponential tail decay, thus selecting $L$ to be, for example, some constant multiplier of $\sigma$, will make sure that outside $\mathcal{A}$ norms are negligible.
We will show the result for a function of two variable and the same result holds for more than two variables, where, for each additional variable we get an additional multiplicative factor of $2L$. Also for simplicity we will assume the box to be $[0,2L]^2$ as opposed to $[-L,-L]^2$. Note that this does not change the analysis.

We have, $\|g\|^2=\int_0^{2L}\int_0^{2L} g^2(v,u_1)dvdu_1$ and $\|g_1\|^2=\int_0^{2L}\left(\int_0^{2L} g(v,u_1)du_1\right)^2dv$. By change of variable, $x=\frac{u_1}{2L}$, we have $\int_0^{2L}g(v,u_1)du_1=2L\int_0^1g(v,2Lx)dx$. Here $dx$ acts as a probability measure and hence applying Jensen's inequality we get
\[\left[\int_0^{2L}g(v,u_1)du_1\right]^2=(2L)^2\left[\int_0^1g(v,2Lx)dx\right]^2\leq(2L)^2\int_0^1g^2(v,2Lx)dx=2L\int_0^{2L}g^2(v,u_1)du_1\]
where the inequality follows from Jensen's and the last equality follows by changing variable one more time. Thus,
\[\|g_1\|^2=\int_0^{2L}\left[\int_0^{2L}g(v,u_1)du_1\right]^2dv\leq\int_0^{2L}\left[2L\int_0^{2L}g^2(v,u_1)du_1\right]dv=2L\|g\|^2\]
For each additional variable we get an additional multiplicative $2L$ term, hence,\\
 $\|g_1\|^2\leq(2L)^{k-1}\|g\|^2\leq(2L)^k\|g\|^2$.
\end{proof}

A version of the following Lemma was proved in \cite{vempala02}. We tailor it for our purpose.

\begin{lemma}\label{lem:svd_projection}
Let the rows of $A \in \mathbb{R}^{N \times n}$  be picked according to a mixture of Gaussians with means $\bs{\mu}_{1},\bs{\mu}_{2},\ldots,$\\$\bs{\mu}_{k}\in \mathbb{R}^{n}$, common variance $\sigma^{2}$ and mixing weights $\alpha_{1}, \alpha_{2},\ldots, \alpha_{k}$ with minimum mixing weight being $\alpha_{\min}$. Let $\tilde{\bs{\mu}}_1,\tilde{\bs{\mu}}_2,\ldots,\tilde{\bs{\mu}}_k$ be the projections of these means on to the subspace spanned by the top $k$ right singular vectors of the sample matrix $A$. Then for any $0<\epsilon<1, 0<\delta<1$,  with probability at least $1-\delta$, $\|\bs{\mu}_{i}-\tilde{\bs{\mu}}_i\|\leq\frac{\epsilon}{2}$, provided 
$N=\Omega\left(\frac{n^{3}\sigma^{4}}{\alpha_{min}^{3}\epsilon^{4}}\left(\log\left(\frac{n\sigma}{\epsilon \alpha_{min}}\right)+\frac{1}{n(n-k)}\log(\frac{1}{\delta})\right)\right)$,
\end{lemma}

\begin{proof}
First note that from Theorem 2 and Corollary 3 of \cite{vempala02}, for $0<\tilde{\epsilon}<\frac{1}{2}$ with probability at least $1-\delta$, we have,
 $\sum_{i=1}^{k}w_{i}(\|\bs{\mu}_{i}\|^{2}-\|\tilde{\bs{\mu}}_i\|^{2})\leq \tilde{\epsilon}(n-k)\sigma^{2}$ provided,
\begin{equation}\label{eq:vempala}
N=\Omega\left(\frac{1}{\tilde{\epsilon}^{2}\alpha_{\min}}\left(n\log\left(\frac{n}{\tilde{\epsilon}}+\max_{i}\frac{\|\bs{\mu}_{i}\|^{2}}{\tilde{\epsilon}\sigma^{2}}\right)+\frac{1}{n-k}\log(\frac{1}{\delta})\right)\right)
\end{equation}

  Now setting $\tilde{\epsilon}=\frac{\hat{\epsilon}^{2}\alpha_{\min}}{n-k}$, we have $\|\bs{\mu}_{i}-\tilde{\bs{\mu}}_i\|^{2}=\|\bs{\mu}_{i}\|^{2}-\|\tilde{\bs{\mu}}_i\|^{2}\leq \hat{\epsilon}^{2}\sigma^{2}$. Next, setting $\epsilon=2\hat{\epsilon}\sigma$ yields the desired result. Note that for this choice  of $\epsilon, \tilde{\epsilon}=\frac{\epsilon^{2} \alpha_{\min}}{4\sigma^{2}(n-k)}$. Further, restricting $0<\epsilon<1$, yields $\tilde{\epsilon}<\frac{w_{\min}}{4\sigma^{2}(n-k)}<\frac{1}{2}$ as required. Now noticing that $\|\bs{\mu}_{i}\|\leq \sqrt{n}$ and plugging in $\tilde{\epsilon}=\frac{\epsilon^{2} \alpha_{\min}}{4\sigma^{2}(n-k)}$ in Equation~\ref{eq:vempala} yields the desired sample size.
\end{proof}

In the following Lemma we consider a mixture of Gaussians where the mixing weights are allowed to take negative values. This might
sound counter intuitive since mixture of Gaussians are never allowed to take negative mixing weights. However, if we have two separate mixtures,
for example, one true mixture density $p(\bs{x})$ and one its estimate $\hat{p}(\bs{x})$, the function $(p-\hat{p})(\bs{x})$ that describes
the difference between the two densities can be thought of as a mixture of Gaussians with negative coefficients. Our goal is to find a bound of the $L^2$ norm of such a function.
\begin{lemma}\label{lem:upper_bound}
Consider a mixture of $m$ k-dimensional Gaussians $f(\bs{x})=\sum_{i=1}^{m}\alpha_i K(\bs{x},\bs{\nu}_{i})$ where the mixing coefficients $\alpha_{i}\in(-1,1),~ i=1,2,\ldots,m$. Then the $L^{2}$ norm of $f$ satisfies $\|f\|^{2}\leq \left(\frac{1}{(2\pi\sigma^{2})^{k}}\right)\bs{\alpha}^{T}\bs{\hat{K}}\bs{\alpha}$, where $\bs{\hat{K}}$ is a $m\times m$ matrix with $\bs{\hat{K}}_{ij}=\exp\left(-\frac{\|\bs{\nu}_{i}-\bs{\nu}_{j}\|^{2}}{2\sigma^{2}}\right)$ and $\bs{\alpha}=(\alpha_{1},\alpha_{2},\ldots,\alpha_{m})^{T}$.
\end{lemma}
\begin{proof}
Let the kernel $t(\bs{x},\bs{y})=\exp\left(-\frac{\|\bs{x}-\bs{y}\|^{2}}{2\sigma^{2}}\right)$ defines a unique RKHS $\mathcal{H}$. Then\\ $\|f\|^{2}_{\mathcal{H}}=\langle \sum_{i=1}^{m}\frac{1}{(\sqrt{2\pi}\sigma)^{k}}\exp\left(-\frac{\|\bs{x}-\bs{\nu}_{i}\|^{2}}{2\sigma^{2}}\right), \sum_{i=1}^{m}\frac{1}{(\sqrt{2\pi}\sigma)^{k}}\exp\left(-\frac{\|\bs{x}-\bs{\nu}_{i}\|^{2}}{2\sigma^{2}}\right)\rangle_{\mathcal{H}}$\\
$=\left(\frac{1}{(2\pi\sigma^{2})^{k}}\right)\langle\sum_{i=1}^{m}\alpha_{i}t(\bs{\nu}_{i}, .), \sum_{i=1}^{m}\alpha_{i}t(\bs{\nu}_{i}, .)\rangle_{\mathcal{H}}$\\
$=\left(\frac{1}{(2\pi\sigma^{2})^{k}}\right)\left\{\sum_{i=1}^{m}\alpha_{i}^{2}+\sum_{i,j,i\neq j}\alpha_{i}\alpha_{j}\langle t(\bs{\nu}_{i}, .), t(\bs{\nu}_{j}, .)\rangle_{\mathcal{H}}\right\}$\\
$=\left(\frac{1}{(2\pi\sigma^{2})^{k}}\right)\left\{\sum_{i=1}^{m}\alpha_{i}^{2}+\sum_{i,j,i\neq j}\alpha_{i}\alpha_{j}t(\bs{\nu}_{i},\bs{\nu}_{j})\right\}=\left(\frac{1}{(2\pi\sigma^{2})^{k}}\right)\bs{\alpha}^{t}\bs{\hat{K}}\bs{\alpha}$.

Since $L_{2}$ norm is bounded by RKHS norm the result follows.
\end{proof}

\noindent\textbf{Proof of Corollary \ref{cor:mixing_weight}}
\begin{proof}
Consider a new mixture $p_o(\bs{x},\bs{m},\bs{\alpha}^*)$ obtained by perturbing the means of $p_o(\bs{x},\bs{m}^*,\bs{\alpha}^*)$.
For ease of notation we use the following short hands $p_o=p_o(\bs{x},\bs{m},\bs{\alpha}), \hat{p}_o=p_o(\bs{x},\bs{m},\bs{\alpha}^*)$
 and $p^*_o=p_o(\bs{x},\bs{m}^*,\bs{\alpha}^*)$. Note that the function $f_1$ mentioned in Theorem \ref{thm:main_theorem} which provides
  the lower bound of a mixture norm, is also a function of $k$ and $\alpha_{\min}$. We will explicitly use this fact here.
Now,\\
$\|p_o-\hat{p}_o\|\leq\|p_o-p^*_o\|+\|p^*_o-\hat{p}_o\|$\\
$~~~~~~~~~~~~~\stackrel{a}{\leq}2\|p_o-p^*_o\|\leq2\left(\|p_{kde}-p^*_o\|+\|p_{kde}-p_o\|\right)$\\
$~~~~~~~~~~~~~\stackrel{b}{\leq}2\left(f_2(G)+\epsilon_*+\epsilon_*\right)$\\
$~~~~~~~~~~~~~\stackrel{c}{\leq}2f_1\left(\frac{\epsilon}{2}\right)=2f_1\left(2k,\alpha_{\min},\frac{\epsilon}{2}\right)$\\
where in equality a follows from the fact $\|p^*_o-\hat{p}_o\|\leq\|p_o-p^*_o\|$ dictated by the upper bound of
Lemma \ref{thm:multivariate_upper_bound}, inequality b follows from Equation \ref{eq:kde_error} and \ref{eq:proxy_error} and
finally inequality c follows from Equation \ref{eq:choice_G}.

It is easy to see that $f_1(k,\beta_{\max},d_{\min}/2)\leq\|p_o-\hat{p}_o\|$ where $\beta_{\max}=\max_{i}\{|\alpha_i-\tilde{\alpha}_i|\}$.
 In order to see this, note that $p_o-\hat{p}_o$ is a mixture of $k$ Gaussians with mixing weights $(\alpha_i-\tilde{\alpha}_i)$ and
 minimum distance between any pair of means is at least $\frac{d_{\min}}{2}$. This is because after projection onto SVD space
 each mean can move by a distance of at most $\frac{\epsilon}{2}$. Thus, minimum pairwise distance between any pairs of projected
 means is at least $d_{\min}-\epsilon\geq\frac{d_{\min}}{2}$ since $\epsilon\leq\frac{d_{\min}}{2}$. Now, choose the
 Gaussian component that has absolute
 value of the mixing coefficient $\beta_{\max}$ and apply the same argument as in Theorem \ref{thm:multivariate_lower_bound}
 (Note that in Lemma \ref{lem:vandermonde_function_bound} we do not need to replace $\beta_{\max}$ by $\beta_{\min}$).

Combining lower and upper bounds we get $f_1(k,\beta_{\max},\frac{d_{\min}}{2})\leq\|p_o-\hat{p}_o\|\leq2f_1(2k,\alpha_{\min},\frac{\epsilon}{2})$.
Simplifying the inequality $f_1(k,\beta_{\max},\frac{d_{\min}}{2})\leq2f_1(2k,\alpha_{\min},\frac{\epsilon}{4})$ and solving for $\beta_{\max}$ yields\\
$\beta_{\max}=\max_{i}|\alpha_i-\tilde{\alpha}_i|\leq\frac{\alpha_{\min}^2}{\sigma^{1/4}}\left(\frac{\epsilon^3}{256n^3k^6}\right)^{C_{2}k}$ for some positive $C_2$ independent of $n$ and $k$. Clearly, $\beta_{\max}\leq\epsilon$.
\end{proof}

\section{Finite Sample Bound for Kernel Density Estimates in High Dimension}
Most of the available literature in kernel density estimate in high dimension provide asymptotic
mean integrated square error approximations, see for example \cite{wand95}, while it is not very
difficult to find an upper bound for the mean integrated square error (MISE) as we will show in
this section. Our goal is to show that for a random sample of sufficiently large size, the integrated
square error based on this sample is close to its expectation (MISE) with high probability.

We will start with a few standard tools that we will require to derive our result.

\noindent\textbf{Multivariate version of Taylor series:}\\
Consider the standard Taylor series expansion with remainder term of a twice differentiable
function $f:\mathbb{R}\rightarrow \mathbb{R}$,
\[f(t)=f(0)+tf'(0)+\int_0^t(t-s)f''(s)ds\]
By change of variable $s=t\tau$ we have the form
\[f(t)=f(0)+tf'(0)+t^2\int_0^\tau(1-\tau)f''(t\tau)d\tau\]
Now a consider a function $g:\mathbb{R}^d\rightarrow \mathbb{R}$ with continuous second order partial
derivatives. For any $\bs{x},\bs{a}\in\mathbb{R}^d$ in the domain of $g$, if we want to expand
$g(\bs{x}+\bs{a})$ around $\bs{x}$, we simply use $u(t)=\bs{x}+t\bs{a}$ and use the one dimensional Taylor series version for the function $f(t)=g(u(t))$. This leads to,
\begin{equation}\label{eq:taylor}
g(\bs{x}+\bs{a})=g(\bs{x})+\bs{a}^T\nabla g(\bs{x})+\int_0^1(1-\tau)\left(\bs{a}^T\mathcal{H}_g(\bs{x}+\tau\bs{a})\bs{a}\right)d\tau
\end{equation}
where $\mathcal{H}_g$ is Hessian matrix of $g$.

\noindent\textbf{Generalized Minkowski inequality}, see \cite{tsybakov09}:\\
For a Borel function $g$ on $\mathbb{R}^d\times \mathbb{R}^d$, we have
\[\int\left(\int g(\bs{x},\bs{y})d\bs{x}\right)^2 d\bs{y}\leq\left[\int\left(\int g^2(\bs{x},\bs{y})d\bs{y}\right)^{1/2} d\bs{x}\right]^2\]

\begin{defi}
Let $L>0$. The Sobolev class $\mathcal{S}(2,L)$  is defined as the set of all functions
$f: \mathbb{R}^d\rightarrow \mathbb{R}$ such that $f\in W^{2,2}$, and all the second order partial derivatives $\frac{\partial^2 f}{\partial x_1^{\alpha_1}\ldots\partial x_d^{\alpha_d}}$, where $\alpha=(\alpha_1,\alpha_2,\ldots,\alpha_d)$ is a multi-index with $|\alpha|=2$ , satisfy
\[\left\|\frac{\partial^2 f}{\partial x_1^{\alpha_1}\ldots\partial x_d^{\alpha_d}}\right\|_2\leq L\]
\end{defi}
Let $\mathcal{H}_f(\bs{x})$ be the Hessian matrix of $f$ evaluated at $\bs{x}$. For any $f\in\mathcal{S}(2,L)$, using Holder's inequality it can  be shown that for any $\bs{a}\in\mathbb{R}^d, \int\left(\bs{a}^T\mathcal{H}_f(\bs{x})\bs{a}\right)^2d\bs{x}\leq L^2\left(\bs{a}^T\bs{a}\right)^2$.
Note that mixture of Gaussians belongs to any Sobolev class.

Given a sample $S=\{X_1, X_2,\ldots,X_N\}$ the kernel density estimator $\hat{p}_S(\cdot)$
of true density $p(\cdot)\in\mathcal{S}(2,L)$ is given by
\begin{equation}
\hat{p}_S(\bs{x})=\frac{1}{Nh^d}\sum_{i=1}^N K\left(\frac{\bs{x}-X_i}{h}\right)
\end{equation}
where $K:\mathbb{R}^d\rightarrow\mathbb{R}$ is a kernel\footnote{Note that normally kernel is a function of two variables i.e., $K:\mathbb{R}^d\times\mathbb{R}^d\rightarrow\mathbb{R}$. However, in nonparametric density estimation literature a kernel function is defined as $\tilde{K}:\mathbb{R}^d\rightarrow\mathbb{R}$, where $K(\bs{x},\bs{y})=\tilde{K}(\bs{x}-\bs{y})$. To be consistent with nonparametric density estimation literature, we will call $\tilde{K}$ as our kernel function and denote it by $K$.} function satisfying\footnote{Note that kernel $K$ here is different from the one introduced in Section \ref{sec:outline}} $\int K(\bs{x})d\bs{x}=1, \int \bs{x}K(\bs{x})d\bs{x}=\bs{0}$ and
$\int \bs{x}^T\bs{x} K(\bs{x})d\bs{x}<\infty$. In particular assume $\int \bs{x}^T\bs{x} K(\bs{x})d\bs{x}\leq C_1$ for some $C_1>0$.
Also let $\int K^2(\bs{x})d\bs{x}\leq C_2$ for some $C_2>0$.
Since the sample $S$ is random, the quantity $\hat{p}_S(\bs{x})$ and
$A_s(X_1, X_2,\ldots,X_N)=\int\left[\hat{p}_S(\bs{x})-p(\bs{x})\right]^2d\bs{x}$, which is square of
the $L^2$ distance between the estimated density and the true density, are also random.
Note that the expected value of $A_S$, $\mathbb{E}(A_s)=\mathbb{E}\int\left[\hat{p}_S(\bs{x})-p(\bs{x})\right]^2d\bs{x}$
is the mean integrated square error (MISE). We will show that for sufficiently large sample
size, $A_s$ is close $\mathbb{E}(A_s)$ with high probability.

First fix any $\bs{x}_0$. The mean square error (MSE) at point $\bs{x}_0$,
$MSE(\bs{x}_0)\stackrel{\Delta}{=}\mathbb{E}\left[(\hat{p}_S(\bs{x})-p(\bs{x}))^2\right]$, where the
expectation is taken with respect to the distribution of $S=(X_1,X_2,\ldots,X_N)$ can be
broken down in to bias and variance part as follows,
$MSE(\bs{x}_0)=b^2(\bs{x}_0)+var(\bs{x}_0)$ where $b(\bs{x}_0)=\mathbb{E}(\hat{p}_S(\bs{x}_0))-p(\bs{x}_0)$ and
$var(\bs{x}_0)=\mathbb{E}\left[\left(\hat{p}_S(\bs{x}_0)-\mathbb{E}[\hat{p}_S(\bs{x}_0)]\right)^2\right]$.

Let us deal with the bias term first. By introducing the notation $K_{\bs{H}}(\bs{u})=|\bs{H}|^{-1/2}K(\bs{H}^{-1/2}\bs{x})$
where $\bs{H}=h^2\bs{I}$, $\bs{I}$ is a $d\times d$ identity matrix and $h>0$ is the kernel bandwidth along all $d$ directions,
we can write
$\hat{p}_S(\bs{x})=\frac{1}{Nh^d}\sum_{i=1}^N K\left(\frac{\bs{x}-X_i}{h}\right)=\frac{1}{Nh^d}\sum_{i=1}^N K\left(\bs{H}^{-1/2}(\bs{x}-X_i)\right)=\frac{1}{N}\sum_{i=1}^N K_H(\bs{x}-X_i)$

Now, $\mathbb{E}(\hat{p}_S(\bs{x}_0))=\mathbb{E} K_{\bs{H}}(\bs{x}_0-X)=\int K_{\bs{H}}(\bs{x}_0-\bs{y})p(\bs{y})d\bs{y}=\int K(\bs{z})p(\bs{x}_0-\bs{H}^{1/2}\bs{z})d\bs{z}$,
where the last inequality follows by change of variables. Expanding $p(\bs{x}_0-\bs{H}^{1/2}z)$ in a Taylor
series around $\bs{x}_0$, using Equation \ref{eq:taylor} we obtain
\[p(\bs{x}_0-\bs{H}^{1/2}\bs{z})=p(\bs{x}_0)-\left(\bs{H}^{1/2}\bs{z}\right)^T\nabla p(\bs{x}_0)+\int_0^1 (1-\tau)\left(\left(\bs{H}^{1/2}\bs{z}\right)^T\mathcal{H}_p(\bs{x}_0-\tau\bs{H}^{1/2}\bs{z})\bs{H}^{1/2}\bs{z}\right)d\tau\]
Thus using $\int K(\bs{z})d\bs{z}=1$ and $\int \bs{z}K(\bs {z})d\bs{z}=\bs{0}$ leads to
\[\mathbb{E}(\hat{p}_S(\bs{x}_0))=p(\bs{x}_0)+h^2\int K(\bs{z})\left[\int_0^1 (1-\tau)\bs{z}^T\mathcal{H}_p(\bs{x}_0-\tau\bs{H}^{1/2}\bs{z})\bs{z}d\tau\right]d\bs{z}\]
 i.e., $b(\bs{x}_0)=\mathbb{E}(\hat{p}_S(\bs{x}_0))-p(\bs{x}_0)=h^2\int K(\bs{z})\left[\int_0^1 (1-\tau)\bs{z}^T\mathcal{H}_p(\bs{x}_0-\tau\bs{H}^{1/2}\bs{z})\bs{z}d\tau\right]d\bs{z}$.

Now,
\[\int b^2(\bs{x})d\bs{x}=\int\left(h^2\int K(\bs{z})\left[\underbrace{\int_0^1(1-\tau)\bs{z}^T\mathcal{H}_p(\bs{x}-\tau\bs{H}^{1/2}\bs{z})\bs{z}d\tau}_{g(\bs{x},\bs{z})}\right]d\bs{z}\right)^2d\bs{x}\]
\[=h^4\int\left(\int K(\bs{z})g(\bs{x},\bs{z})d\bs{z}\right)^2d\bs{x}\leq h^4\left[\int\left(\int K^2(\bs{z})g^2(\bs{x},\bs{z})dx\right)^{1/2}d\bs{z}\right]^2\]
\[=h^4\left[\int\left(\int K^2(\bs{z})\left[\int_0^1(1-\tau)\bs{z}^T\mathcal{H}_p(\bs{x}-\tau\bs{H}^{1/2}\bs{z})\bs{z}d\tau\right]^2d\bs{x}\right)^{1/2}d\bs{z}\right]^2\]
\[\leq h^4\left[\int K(\bs{z})\left(\int_0^1\left[\int\left[(1-\tau)\bs{z}^T\mathcal{H}_p(\bs{x}-\tau\bs{H}^{1/2}\bs{z})\bs{z}\right]^2 d\bs{x}\right]^{1/2}d\tau\right) dz\right]^2\]
\[\leq h^4\left[\int K(\bs{z})\left(\int_0^1(1-\tau)L\bs{z}^T\bs{z}d\tau\right)d\bs{z}\right]^2=\frac{L^2h^4}{4}\left(\int \bs{z}^T\bs{z}K(\bs{z})dz\right)^2\leq \frac{C_1^2L^2}{4}h^4\]
where the first and second inequality follows by applying Generalized Minkowski inequality. The third inequality
follows from the fact that $p\in Sob(2,L)$ and support of $p$ is the whole real line.

Now let us deal with the variance term.
Let $\eta_i(\bs{x}_0)=K\left(\frac{\bs{x}_0-X_i}{h}\right)-\mathbb{E}\left[K\left(\frac{\bs{x}_0-X_i}{h}\right)\right]$.
The random variables $\eta_i(\bs{x}_0), i=1,\ldots,N$ are iid with zero mean and variance
\[\mathbb{E}\left[\eta_i^2(\bs{x}_0)\right]\leq \mathbb{E}\left[K^2\left(\frac{\bs{x}_0-X_i}{h}\right)\right]=\int K^2\left(\frac{\bs{x}_0-\bs{z}}{h}\right)p(\bs{z})d\bs{z}\]
Then,
\[var(\bs{x}_0)=\mathbb{E}\left[\left(\hat{p}_S(\bs{x}_0)-\mathbb{E}[\hat{p}_S(\bs{x}_0)]\right)^2\right]=\mathbb{E}\left[\left(\frac{1}{Nh^d}\sum_{i=1}^N \eta_i(\bs{x}_0)\right)^2\right]\]
\[=\frac{1}{Nh^{2d}}\mathbb{E}\left[\eta_1^2(\bs{x}_0)\right]\leq\frac{1}{Nh^{2d}}\int K^2\left(\frac{\bs{x}_0-\bs{z}}{h}\right)p(\bs{z})d\bs{z}\]
Clearly.
\[\int var(\bs{x})dx\leq\frac{1}{Nh^{2d}}\int\left[\int K^2\left(\frac{\bs{x}-\bs{z}}{h}\right)p(\bs{z})d\bs{z}\right]d\bs{x}=\frac{1}{Nh^{2d}}\int p(\bs{z})\left[\int K^2\left(\frac{\bs{x}-\bs{z}}{h}\right)d\bs{x}\right]d\bs{z}\]
\[=\frac{1}{Nh^d}\int K^2(\bs{v})d\bs{v}\leq \frac{C_2}{Nh^d}\]
Now,
\[MISE=\mathbb{E}(A_s)=\mathbb{E}\int\left[\hat{p}_S(\bs{x})-p(\bs{x})\right]^2d\bs{x}=\int\mathbb{E}\left[\hat{p}_S(\bs{x})-p(\bs{x})\right]^2d\bs{x}=\int MSE(\bs{x})d\bs{x}\]
\[=\int b^2(\bs{x})d\bs{x}+ \int var(\bs{x})d\bs{x}\leq \frac{C_1^2L^2}{4}h^4+\frac{C_2}{Nh^d}\]
The bias and variance terms can be balanced by selecting $h^*=\left(\frac{C_2}{C_1^2L^2}\right)^{\frac{1}{d+4}}\left(\frac{d}{N}\right)^{\frac{1}{d+4}}$.
With this choice of $h$ we have $MISE\leq \frac{4+d}{4d}\left(\frac{(C_1^2L^2)^dC_2^dd^4}{N^4}\right)^{\frac{1}{d+4}}$.
Note that this is of the order $N^{-\frac{4}{d+4}}$. Similar expressions for bias/variance terms and convergence
rate are also known to hold, but with different constants, for asymptotic MISE approximations (see \cite{wand95}).

Since mixture of Gaussians belongs to any Sobolev class, the following Lemma shows that we can approximate
the density of such a mixture arbitrarily well in $L^2$ norm sense.
\begin{lemma}\label{lem:kde_estimate}
Let $p\in \mathcal{S}(2,L)$ be a $d$-dimensional probability density function and $K:\mathbb{R}^d\rightarrow\mathbb{R}$ be any kernel function with diagonal bandwidth matrix $h^2\bs{I}$, satisfying
$\int K(\bs{x})d\bs{x}=1, \int \bs{x}K(\bs{x})d\bs{x}=\bs{0}, \int \bs{x}^T\bs{x}K(\bs{x})d\bs{x}<C_1$ and $\int K^2(\bs{x})d\bs{x}<C_2$ for positive $C_1, C_2$. Then for any
$\epsilon_0>0$ and any $\delta\in(0,1)$, with probability grater than $1-\delta$, the kernel density
estimate $\hat{p}_S$ obtained using a sample $S$ of size
$\Omega\left(\left[\frac{\log (1/\delta)}{\epsilon_0^2}\right]^d\right)$ satisfies,
$\int\left(p(\bs{x})-\hat{p}_S(\bs{x})\right)^2d\bs{x}\leq \epsilon_0$.
\end{lemma}
\begin{proof}
For a sample $S=\{X_1,X_2,\ldots,X_N\}$ we will use the notation $A_S=A_S(X_1,X_2,\ldots,X_i,\ldots,X_N)$ to denote the random quantity
$\int\left(p(x)-\hat{p}_S(x)\right)^2dx$. Note that $\mathbb{E}(A_S)=MISE$. Our goal is to use a
large enough sample size so that $A_S$ is close to its expectation. In particular we would like to use
McDiarmid's inequality to show that
\[Pr\left(A_S-\mathbb{E}(A_S)>\frac{\epsilon_0}{2}\right)\leq\exp\left(-\frac{2(\frac{\epsilon_0}{2})^2}{\sum_{i=1}^N c_i^2}\right)\]
where, $\sup_{\bs{x}_1,\ldots,\bs{x}_i,\ldots,\bs{x}_N,\hat{\bs{x}}_i}|A_S(\bs{x}_1,\ldots,\bs{x}_i,\ldots,\bs{x}_N)-A_S(\bs{x}_1,\ldots,\hat{\bs{x}}_i,\ldots,\bs{x}_N)|\leq c_i$ for $1\leq i\leq N$. Let
\[B_i=\int\underbrace{\left(\frac{1}{Nh^d}\left[K\left(\frac{\bs{x}-X_1}{h}\right)+\ldots+K\left(\frac{\bs{x}-X_i}{h}\right)+\ldots+K\left(\frac{\bs{x}-X_N}{h}\right)\right]-p(\bs{x})\right)^2}_{b_i}d\bs{x}\]
\[\hat{B}_i=\int\underbrace{\left(\frac{1}{Nh^d}\left[K\left(\frac{\bs{x}-X_1}{h}\right)+\ldots+K\left(\frac{\bs{x}-\hat{X}_i}{h}\right)+\ldots+K\left(\frac{\bs{x}-X_N}{h}\right)\right]-p(\bs{x})\right)^2}_{\hat{b}_i}d\bs{x}\]
Then,
\[b_i-\hat{b}_i=\frac{1}{N^2h^{2d}}\left[K^2\left(\frac{\bs{x}-X_i}{h}\right)-K^2\left(\frac{\bs{x}-\hat{X}_i}{h}\right)\right]-\frac{2p(\bs{x})}{Nh^d}\left[K\left(\frac{\bs{x}-X_i}{h}\right)-K\left(\frac{\bs{x}-\hat{X}_i}{h}\right)\right]\]
\[+\frac{2}{N^2h^{2d}}\sum_{j\neq i}K\left(\frac{\bs{x}-X_j}{h}\right)\left[K\left(\frac{\bs{x}-X_i}{h}\right)-K\left(\frac{\bs{x}-\hat{X}_i}{h}\right)\right]\]

After integrating, the first term in the above equation can be bounded by $\frac{2C_2}{N^2h^d}$, second 
term can be bounded by $\frac{4\left(\sqrt{C_2\int p^2(\bs{x})d\bs{x}}\right)}{N}$ and the third term can be bounded by $\frac{4C_2}{Nh^d}$.
Thus, $|B_i-\hat{B}_i|\leq \frac{2C_2}{N^2h^d}+\frac{4\left(\sqrt{C_2\int p^2(\bs{x})d\bs{x}}\right)}{N}+\frac{4C_2}{Nh^d}$.

Note that the optimal choice of $h$ of the order $\left(\frac{d}{N}\right)^{\frac{1}{d+4}}$
as derived previously does not help to get a tight concentration inequality type bound. However,
we can choose a suitable $h$ that solve our purpose. To this aim,
we assume that $|B_i-\hat{B}_i|$ is dominated by term $\frac{1}{Nh^d}$, i.e.,
\begin{equation}\label{eq:dominate}
\frac{1}{N}\leq\frac{1}{Nh^d}
\end{equation}
later we need to show that this is indeed satisfied for the choice of $h$. Thus,
\[c_i=|B_i-\hat{B}_i|=\sup_{\bs{x}_1,\ldots,\bs{x}_i,\ldots,\bs{x}_N,\hat{\bs{x}}_i}|A_S(\bs{x}_1,\ldots,\bs{x}_i,\ldots,\bs{x}_N)-A_S(\bs{x}_1,\ldots,\hat{\bs{x}}_i,\ldots,\bs{x}_N)|\leq\frac{C}{Nh^d}\]
where $C$ is a function of $C_1,C_2$ and $L$. Now McDiarmid's inequality yields
\begin{equation}\label{eq:concentration}
Pr\left(A_S-\mathbb{E}(A_S)>\frac{\epsilon_0}{2}\right)\leq\exp\left(-\frac{\epsilon_0^2Nh^{2d}}{2C^2}\right)=\exp\left(-\frac{\epsilon_0^2N^{\beta}}{2C^2}\right)
\end{equation}
where we have set $Nh^{2d}=N^{\beta}$ for some $\beta>0$. Setting right side of equation
\ref{eq:concentration} less
than or equal to $\delta$, we get $\left(\frac{2C^2\log\left(\frac{1}{\delta}\right)}{\epsilon_0^2}\right)^{1/\beta}\leq N$.
Now setting $\beta=1/d$, we get $\left(\frac{2C^2\log\left(\frac{1}{\delta}\right)}{\epsilon_0^2}\right)^d\leq N$.
For this
choice of $\beta$, solving $Nh^{2d}=N^{\beta}$ we get $h=\frac{1}{N^{\frac{d-1}{2d^2}}}$. Now setting this
value of $h$ we get $\frac{1}{Nh^d}=\frac{1}{N^{\frac{1}{2}+\frac{1}{2d}}}$. For $d>1$ this rate is indeed slower
than $\frac{1}{N}$ and hence Equation \ref{eq:dominate} is satisfied. Next we check what is the convergence rate
of MISE for this choice of $h$. Ignoring the constant terms, the bias terms is of the order
$h^4=\frac{1}{N^{\frac{2(d-1)}{d^2}}}$, whereas the variance term is of the order
$\frac{1}{N^d}=\frac{1}{N^{\frac{1}{2}+\frac{1}{2d}}}$. Since the bias term decreases at a much slower rate,
convergence rate of MISE is dominated by the bias term and hence $MISE\leq \frac{C^*}{N^{\frac{2(d-1)}{d^2}}}$ for
some constant $C^*$ independent of $d$ and $N$.
Thus to make sure that $MISE=\mathbb{E}(A_S)\leq\frac{\epsilon_0}{2}$, we need
$\left(\frac{2C^*}{\epsilon_0}\right)^{\frac{d^2}{2(d-1)}}\leq N$. Since
$\left(\frac{2C^*}{\epsilon_0}\right)^{\frac{d^2}{2(d-1)}}\leq \left(\frac{2C^*}{\epsilon_0}\right)^d$, $\left(\frac{2C^*}{\epsilon_0}\right)^d\leq N$ will suffice. However, the
number of examples required $\left(\frac{2C^2\log\left(\frac{1}{\delta}\right)}{\epsilon_0^2}\right)^{d}$
to ensure that with probability greater than $1-\delta$, $A_S\leq \mathbb{E}(A_S)+\frac{\epsilon_0}{2}$ is much
higher than $\left(\frac{2C^*}{\epsilon_0}\right)^d$ and hence for any sample of this size, $\mathbb{E}(A_S)\leq\frac{\epsilon_0}{2}$.
The result follows.
\end{proof}

For the sake of completeness we present McDiarmid's inequality below.
\begin{lemma}
Let $X_1,X_2,\ldots, X_N$ be iid random variables taking values in a set $A$, and assume that
$f:A^N\rightarrow \mathbb{R}$ is a function satisfying
\[\sup_{\bs{x}_1,\bs{x}_2,\ldots,\bs{x}_N,\hat{\bs{x}}_i}|f(\bs{x}_1,\bs{x}_2,\ldots,\bs{x}_N)-f(\bs{x}_1,\bs{x}_2,\ldots,\bs{x}_{i-1},\hat{\bs{x}}_i,\bs{x}_{i+1},\ldots,\bs{x}_N)|\leq c_i\]
for $1\leq i\leq N$. Then for any $\epsilon>0$,
\[Pr\left\{f(X_1,X_2,\ldots,X_N)-\mathbb{E}[f(X_1,X_2,\ldots,X_N)]\geq\epsilon\right\}\leq\exp\left(-\frac{2\epsilon^2}{\sum_{i=1}^N c_i^2}\right)\]
\end{lemma}

\section{Estimation of Unknown Variance}\label{app:variance}

We now provide the proof of Theorem \ref{thm:variance_estimation} which combines results from the remainder of this Appendix.\\

\noindent\textbf{Proof of Theorem \ref{thm:variance_estimation}:}
It is shown in Lemma 5A of~\cite{lindsay89} that the smallest 
positive root of the determinant $d(\tau)=\det(M)(\tau)$, viewed is a function of $\tau$, is equal to the variance $\sigma$ of the original mixture $p$ and also that $
d(\tau)$ undergoes a   sign change at its smallest positive root. Let the smallest positive root of $\hat{d}(\tau)= \det(\hat{M})(\tau)$ be $\hat{\sigma}$. We now show for any $\epsilon>0$
that  $\sigma$ and $\hat{\sigma}$ are within $\epsilon$ given $O\left(\frac{n^{\poly(k)}}{\epsilon^2\delta}\right)$  samples.

In Corollary \ref{cor:independent_leading_term} we show that both $d(\tau)$
and $\hat{d}(\tau)$ are polynomials of degree $k(k+1)$ and the highest degree coefficient of $\hat{d}(\tau)$ is independent of the sample.
The rest of the coefficients of $d(\tau)$ and $\hat{d}(\tau)$ are sums of products of the 
coefficients of   individual entries of the matrices $M$ and $\hat{M}$ respectively.

Note that $\mathbb{E}(\hat{M})=M$, i.e., for any $1\leq i,j,\leq (k+1), \mathbb{E}(\hat{M}_{i,j}(\tau))=M_{i,j}(\tau)$. Since $M_{i,j}(\tau)$ is a polynomial in $\tau$, using standard 
concentration results we can show that coefficients of the polynomial $\hat{M}_{i,j}(\tau)$ are close to the corresponding coefficients of the polynomial $M_{i,j}(\tau)$ given large enough sample size. 
Specifically, we show in Lemma \ref{lem:matrix_entry_concentration} that given a sample of size $O\left(\frac{n^{\poly(k)}}{\epsilon^2\delta}\right)$ each of the coefficients of each of the polynomials $M_{i,j}(\tau)$ can be estimated within error $O\left(\frac{\epsilon}{n^{\poly(k)}}\right)$ with probability at least $1-\delta$.

Next,  in 
Lemma \ref{lem:det_coeff_conc} we show that estimating each of the coefficients  of the polynomial $M_{i,j}(\tau)$ for all $i,j$ with accuracy $O\left(\frac{\epsilon}{n^{\poly(k)}}\right)$  
ensures that all coefficients of $d(\hat{\tau})$ are $O\left(\frac{\epsilon}{k}\right)$ close to the corresponding coefficients of ${d}(\tau)$ with high probability.

Consequently, in Lemma \ref{lem:coefficient_and_root} we show that when all coefficients of $\hat{d}(\tau)$ 
are within $O\left(\frac{\epsilon}{k}\right)$  of the corresponding coefficients of ${d}(\tau)$,
the smallest positive root of $\hat{d}(\tau)$, $\hat{\sigma}$, is at most $\epsilon$ away from the smallest positive
root ${\sigma}$ of ${d}(\tau)$.

Observing that there exist  many efficient numerical methods for estimating roots of polynomial of one variable within the desires accuracy completes the proof.

\qed

\begin{lemma}\label{lem:Hankel}
Consider the $(k+1)\times (k+1)$ Hankel matrix $\Gamma$,  $\Gamma_{ij}=(\gamma_{i+j}(x,\tau))$ for $i,j=0,1,...,k$, where
$\gamma_n(x,\tau)$ is the $n^{th}$ Hermite polynomial as described above. Then $\det(\Gamma)(x,\tau)$ is a homogeneous
polynomial of degree $k(k+1)$ of two variables $x$ and $\tau$.
\end{lemma}
\begin{proof}
It is easy to see from the  definition that the $n$th Hermite polynomial $\gamma_n(x,\tau)$ is a homogeneous polynomial  of two variables of degree $n$. Thus we can represent the degree of each polynomial term of the matrix $\Gamma$ as follows
\[\left[\begin{array}{ccccc}
0 & 1 & 2 & \cdots & k\\
1 & 2&  3& \cdots & k+1\\
2 & 3 & 4 & \cdots & k+2\\
\cdots& \cdots& \cdots &\cdots & \cdots\\
k & k+1 & k+2 & \cdots & 2k
\end{array}\right]
\]
Now reduce the degree of each element of row $i$  by taking degree $(i-1)$ by taking it outside the matrix.
The resulting matrix will have degree $(i-1)$ for all the elements in column $i, i=1,2,...,(k+1)$. 
Then reduce the degree of each element of column $i$  by $(i-1)$ by taking it outside the matrix.
The degree of each element of the resulting matrix is $0$.
The remaining matrix has zeros everywhere. Thus we see that when  the determinant is computed, the degree of each (homogenous) term is $2\times(1+2+\cdots+k) = k(k+1)$.
\end{proof}

We have the following simple corollary.
\begin{corollary}\label{cor:independent_leading_term}
$d(\tau)$ is a polynomial of of degree $k(k+1)$, with the coefficient of the leading
term  independent of the probability distribution $p$. Similarly, $\hat{d}(\tau)$
is a polynomial of of degree $k(k+1)$, with the leading term having coefficient
independent of the coefficients of the sampled data.
\end{corollary}
\begin{proof}
From Lemma \ref{lem:Hankel}, notice that $\det(\Gamma(x,\tau))$ is a homogeneous polynomial of degree $k(k+1)$ and hence the non-zero
term $\tau^{k(k+1)}$ cannot include $x$. Since $M(\tau)$ is obtained by replacing $x^i$ by $\mathbb{E}(x^i)$, the leading term of $d(\tau)$ is independent of the probability distribution. Similarly, $\hat{M}(\tau)$ is obtained by replacing $x^i$ by $\frac{\sum_{j=1}^N X_j^i}{N}$ and the result follows.
\end{proof}

\begin{lemma}\label{lem:root_perturbation}
Let $f(x)=x^m+a_{m-1}x^{m-1}+a_{m-2}x^{m-2}+\cdots+a_1x+a_0$ be a polynomial having a smallest positive real root $x_0$ with multiplicity one and $f'(x_0)\neq 0$.  Let $\hat{f}(x)=x^m+\hat{a}_{m-1}x^{m-1}+\hat{a}_{m-2}x^{m-2}+\cdots+\hat{a}_1x+\hat{a}_0$ be another polynomial such that $\|a-\hat{a}\|\leq \epsilon$ for some  sufficiently small $\epsilon>0$, where $a=(a_0,a_1,\ldots,a_{m-1})$ and $\hat{a}=(\hat{a}_0,\hat{a}_1,\ldots,\hat{a}_{m-1})$. Then there exists a $C>0$ such that the smallest positive root $\hat{x}_0$ of $\hat{f}(x)$ satisfies $\|x_0-\hat{x}_0\|\leq C\epsilon$.
\end{lemma}
\begin{proof}
 Let $a=(a_0,a_1,\ldots,a_{m-1})$ be the coefficient vector. The root of the polynomial can be written as a function of the coefficients such that $x(a)=x_0$. Thus we have $x^{m}(a)+a_{m-1}x^{m-1}(a)+a_{m-2}x^{m-2}(a)+\cdots+a_1x(a)+a_0=0$. Taking partial derivative with respect to $a_i$ we have,
$$\frac{\partial x(a)}{\partial a_i}\left[mx^{m-1}(a)+a_{m-1}(m-1)x^{m-2}(a)+a_{m-2}(m-2)x^{m-3}(a)+\cdots+a_22x(a)+a_1\right]+x^{i}(a)=0$$
so that we can write

$$
\|\nabla x(a)\| =\frac{\sqrt{\sum_{i=0}^{m-1}x^{2i}(a)}}{|f'(x(a))|}
$$

Note that $|f'(x)|$ at the root $x=x_0$ is  lower bounded by some $c_1>0$. Since $f''(x)$ is also a polynomial, $|f''(x)|$ can be upper bounded by another $c_2>0$  within a small neighborhood of $x_0$ and hence $|f'(x)|$ can be lower bounded by some $c_3>0$ within the small neighborhood around $x_0$. This neighborhood can also be specified by all $\xi$ within a ball $\mathcal{B}(a,\epsilon)$ of radius $\epsilon>0$, sufficiently small, around $a$. For  sufficiently small $\epsilon$, the polynomial $\sum_{i=0}^{m-1}x^i(\xi)$, where $\xi\in\mathcal{B}(a,\epsilon)$, must be upper bounded by some $c_4>0$. Thus there exists some constant $C>0$ such that $\sup_{\xi\in\mathcal{B}(a,\epsilon)}\|\nabla x(\xi)\|\leq C$.

Now applying mean value theorem,
$$
|x(a)-x(\hat{a})|\leq\|a-\hat{a}\|\sup_{\xi\in\mathcal{B}(a,\epsilon_4)}\|\nabla x(\xi)\|\leq C\epsilon
$$
\end{proof}

\begin{lemma}\label{lem:coefficient_and_root}
Let $\sigma$ be the smallest positive root of $d(\tau)$. Suppose $\hat{d}(\tau)$  be the polynomial
where each of the coefficients of $d(\tau)$ are estimated within $\epsilon$ error for some sufficiently small $\epsilon>0$. Let
$\hat{\sigma}$ be the smallest positive root of  $\hat{d}(\tau)$. Then $|\hat{\sigma}-\sigma|=O(k\epsilon)$.
\end{lemma}
\begin{proof}
We have shown in Corollary \ref{cor:independent_leading_term} that $d(\tau)$ is a polynomial of degree $k(k+1)$ and the leading term has some constant coefficient. Consider a fixed set of $k$ means. This fixed set of means will give rise to a polynomial $d(\tau)$ and $\hat{d}(\tau)$, where means contribute in deciding the coefficients of the corresponding polynomials, for which according to Lemma~\ref{lem:root_perturbation}, there exists a $C>0$ such that $|\hat{\sigma}-\sigma|\leq Ck\epsilon$. Since all possible sets of $k$ means form a compact subset, there exists a positive minimum of all the $C$s. Let this minimum be $C^*$. This proves that $|\hat{\sigma}-\sigma|=O(k\epsilon)$.
\end{proof}

\subsection{Properties of the entries of matrix $M$}
From the construction of the matrix $M$ it is clear that it has $2k$ different entries. Each such entry is a polynomial in $\tau$. Let us denote these distinct entries by $m_i(\tau)=\mathbb{E}[\gamma_i(x,\tau)], i=1,2,\ldots,2k$. Due to the recurrence relation of the Hermite polynomials we observe the following properties of $m_i(\tau)s$,
\begin{enumerate}
\item If $i$ is even then maximum degree of the polynomial $m_i(\tau)$ is $i$ and if $i$ is odd then maximum degree is $(i-1)$.
\item For any $m_i(\tau)$, each term of $m_i(\tau)$ has an even degree of $\tau$. Thus each $m_i(\tau)$ can have at most $i$ terms.
\item The coefficient of each term of $m_i(\tau)$ is multiplication of a constant and an expectation.The constant can be at most $(2k)!$ and the expectation can be, in the worst case, of the quantity $X^{2k}$, where $X$ is sampled from $p$.
\end{enumerate}
Note that the empirical version of the matrix $M$ is $\hat{M}$ where each entry $m_i(\tau)$ is replaced by its empirical counterpart $\hat{m}_i(\tau)$.
Using standard concentration inequality we show that for any $m_i(\tau)$, its coefficients are arbitrarily close to the corresponding coefficients of $\hat{m}_i(\tau)$ provided a large enough sample size is used to estimate $\hat{m}_i(\tau)$.
\begin{lemma}\label{lem:matrix_entry_concentration}
For any $m_i(\tau), i=1,,2,\ldots,2k$, let $\beta$ be any arbitrary coefficient of the polynomial $m_i(\tau)$. Suppose $X_1,X_2,\ldots,X_N$ iid samples from $p$ is used to estimate $\hat{m}_i(\tau)$ and the corresponding coefficient is $\hat{\beta}$. Then there exists a polynomial $\eta_1(k)$ such that for any $\epsilon>0$ and $0<\delta, |\beta-\hat{\beta}|\leq\epsilon$ with probability at least $1-\delta$, provided $N>\frac{n^{\eta_1(k)}}{\epsilon^2\delta}$.
\end{lemma}

\begin{proof}
Note that in the worst case $\beta$ may be a multiplication of a constant which can be at most $(2k)!$ and the quantity $\mathbb{E}(X^{2p})$.
First note that $\mathbb{E}\left(\frac{1}{N}\sum_{i=1}^NX_i^{2k}\right)=\mathbb{E}(X^{2k})$.
Now, 
\begin{eqnarray*}
 \var\left(\frac{1}{N}\sum_{i=1}^NX_i^{2k}\right) &=& \frac{\var(X^{2k})}{N}\\
&=&\frac{1}{N}\mathbb{E}\left(X^{2k}-\mathbb{E}(X^{2k})\right)^2\\
&=& \frac{1}{N}\left(\mathbb{E}(X^{4k})-\left(\mathbb{E}(X^{2k})\right)^2\right)\\
&\leq&\frac{1}{N}\mathbb{E}(X^{4k})\\
&\leq& \frac{(16nk^2)^{2k}}{N}\\
\end{eqnarray*}

The last inequality requires a few technical things. First note that once the Gaussian mixture is projected from $\mathbb{R}^n$ to $\mathbb{R}$ mean of each component Gaussian lies within the interval $[-\sqrt{n},\sqrt{n}]$.
Next note that for any $X\sim\mathcal{N}(\mu,\sigma^2)$, expectation of the quantity $X^i$ for any $i$ can be given by the recurrence relation $\mathbb{E}(X^i)=\mu\mathbb{E}(X^{i-1})+(i-1)\sigma^2\mathbb{E}(X^{i-2})$. From this recurrence relation we see that $\mathbb{E}(X^{4k}$ is a homogeneous polynomial of degree $4k$ in $\mu$ and $\sigma$. Since $|\mu|\leq\sqrt{n}$ and assuming $\sigma\leq\sqrt{n}$ each term of this homogeneous polynomial is less than $(\sqrt{n})^{4k}=n^{2k}$. Next we argue that the homogeneous polynomial $\mathbb{E}(X^{4k})$ can have at most $(4k)!$ terms. To see this  let $x_i$ be the sum of the coefficients of the terms in appearing in the homogeneous polynomial representing expectation of $X^i$.  Note that $x_0=x_1=1$. And for $i\geq 2,~~ x_i=x_{i-1}+(i-1)x_{i-2}$. Using this
recurrence relation, we have $x_{4p}=x_{4p-1}+(4p-1)x_{4p-2}\leq x_{4p-1}+(4p-1)x_{4p-1}=4px_{4p-1}\leq 4p(4p-1)x_{4p-2}\leq 4p(4p-1)(4p-2)x_{4p-3}=\cdots=4p(4p-1)(4p-2)(4p-3)
\cdots (3)(2)(1)=(4p)!$. Thus the homogeneous polynomial representing expectation of $X^{4k}$ has at most $(4k)!$ terms and each term is at most $n^{2k}$. This ensures that $\mathbb{E}(X^{4k})\leq (4k)!n^{2k}\leq (4k)^{4k}n^{2k}\leq(16nk^2)^{2k}$. Note that this upper bound also holds when $X$ is samples from a mixture of $k$ univariate Gaussians.


Now applying Chebyshev's inequality, we get \\
$P\left(\left| \frac{1}{N}\sum_{i=1}^NX_i^{2k}-\mathbb{E}(X^{2k})\right|>\frac{\epsilon}{(2k)!}\right)\leq\frac{\left((2k)!\right)^2\var\left(\frac{1}{N}\sum_{i=1}^NX_i^k\right)}{\epsilon^2}\leq\frac{(2k)^{4k}(16nk^2)^{2k}}{N\epsilon^2}\leq\frac{(64nk^4)^{2k}}{N\epsilon^2}$.

Noting that the constant term in $\beta$ can be at most $(2k)!$ and upper bounding the last quantity above by $\frac{\delta}{2k}$ and applying union bound ensures the existence of a polynomial $\eta_1(k)$ and yields the desired result.
\end{proof}

\subsection{Concentration of coefficients of $d(\tau)$}
In this section we show that if the coefficients of the individual entries  of the matrix $M$ (recall each such entry is a polynomial of $\tau$) are estimated arbitrarily well then the coefficients of $d(\tau)$ are also estimated arbitrarily well.
\begin{lemma}\label{lem:det_coeff_conc}
There exists a polynomial $\eta_2(k)$ such that if coefficients of each of the entries of matrix $M$ (where each such entry  is a polynomial of $\tau$) are estimated within error $\frac{\epsilon}{n^{\eta_2(k)}}$ then each of the coefficients of $d(\tau)$ are estimated within $\epsilon$ error.
\end{lemma}
\begin{proof}
First note that $M$ is a $(k+1)\times (k+1)$ matrix. While computing the determinant, each entry of the matrix $M$ is multiplied to $k$ different entries of the matrix. Further each entry of the matrix (which is a polynomial in $\tau$) can have at most $2k$ terms. Thus in the determinant $d(\tau)$, each of the coefficients of $\tau^{2i}, i=1,2,\ldots,\frac{k(k+1)}{2}$ has only $\eta_4(k)$ term for some polynomial $\eta_4(k)$. Consider any one of the $\eta_4(k)$ terms and let us denote it by $b$. Note that $b$ is multiplication of at most $k$ coefficients of the entries of $M$. Without loss of generality let us denote $b=\beta_i\beta_2\ldots\beta_l$ where $l$ can be at most $k$. Let $\hat{b}$ be the estimation of $b$ given by $\hat{b}=\hat{\beta}_1\hat{\beta}_2\ldots\hat{\beta}_l$ such that for any $1\leq i\leq l, |\beta_i-\hat{\beta}_i|\leq\epsilon_*$ for some $\epsilon_*>0$. For convenience we will write $\hat{\beta}_i=\beta_i+\epsilon^*$. Then we can write
\begin{eqnarray*}
|b-\hat{b}|&=& |\beta_1\beta_2\ldots\beta_l-(\beta_1+\epsilon_*)(\beta_2+\epsilon_*)\ldots(\beta_l+\epsilon_*)|\\
&\leq&(a_1\epsilon_*+a_2\epsilon_*^2+\cdots+a_{l-1}\epsilon_*^{(l-1)}+\epsilon_*^l)\\
&\leq& (a_1+a_2+\cdots+a_{l-1}+1)\epsilon_*\\
\end{eqnarray*}
where $a_i$ is a summation of $\eta_3(k)$ terms for some polynomial $\eta_3(k)$ and each term is a multiplication of at most $(l-1), \beta_j$s. Note that each $\beta_j$ can have value at most $(2k)!(2k)!(\sqrt{n})^{2k}\leq (2k)^{4k}n^k=(16nk^4)^k$. Thus $(a_1+a_2+\cdots+a_{l-1}+1)\leq k\eta_3(k)(16nk^4)^k$. Clearly $|b-\hat{b}|\leq k\eta_3(k)(16nk^4)^k\epsilon_*$. Thus there exists some polynomial $\eta_2$ such that if we set $\epsilon_*=\frac{\epsilon_2}{n^{\eta_2(k)}}$,  then the coefficients of of $d(\tau)$ are estimated  within error $k\eta_3(k)(16nk^4)^k\frac{\epsilon}{n^{\eta_2(k)}}\leq\epsilon$.
\end{proof}
\end{document}